\documentclass[9pt,twoside]{extarticle}
\usepackage{times}
\title{Online learning with kernel losses}
\usepackage[papersize={17cm,24cm},margin=22mm,top=17mm,headsep=5mm]{geometry} 
\usepackage{amsmath}
\usepackage{amsthm}
\usepackage{amssymb}
\usepackage{csquotes} 
\usepackage{graphicx} 
\usepackage[linesnumbered,ruled]{algorithm2e}
\usepackage{thmtools}
\usepackage{thm-restate}
\usepackage[dvipsnames]{xcolor}
\usepackage[english]{babel} 
\usepackage[colorlinks,linkcolor = Aquamarine, urlcolor  = violet, citecolor = YellowOrange, anchorcolor = violet]{hyperref}
\usepackage[english]{babel} 
\date{}
\renewenvironment{abstract}
  {\noindent\small 
  {\noindent{\large\textbf\abstractname. }
  \thispagestyle{plain}
  }}
  {}

\usepackage{fancyhdr}
\renewcommand{\headrulewidth}{0.4pt}
\fancypagestyle{plain}{
  \fancyhead[C,R]{}
  \renewcommand{\headrulewidth}{0pt} 
  \fancyfoot{} 
  }

\usepackage{natbib}
\newcommand\numberthis{\addtocounter{equation}{1}\tag{\theequation}}

\def\lv{\lVert}
\def\rv{\rVert}
\def\ke{\mathcal{K}}
\def\kc{\hat{\mathcal{K}}_m}

\newcommand*\diff{\mathop{}\!\mathrm{d}}

\newcommand{\BlackBox}{\rule{1.5ex}{1.5ex}}  
\newenvironment{Proof}{\par\noindent{\bf Proof\ }}{\hfill\BlackBox\\[2mm]}
\newtheorem{theorem}{Theorem}
\newtheorem{lemma}[theorem]{Lemma}
\newtheorem{proposition}[theorem]{Proposition}
\newtheorem{remark}[theorem]{Remark}
\newtheorem{corollary}[theorem]{Corollary}
\newtheorem{definition}[theorem]{Definition}
\newcommand{\argmin}{\operatornamewithlimits{argmin}}
\newcommand*\samethanks[1][\value{footnote}]{\footnotemark[#1]}

\author{Aldo Pacchiano  \thanks{A. Pacchiano and N. S. Chatterji contributed equally to this work.} \thanks{pacchiano@berkeley.edu; Computer Science Division, UC Berkeley.} \and 
  Niladri S. Chatterji \samethanks[1] \thanks{niladri.chatterji@berkeley.edu; Department of Physics, UC Berkeley.} \and Peter L. Bartlett \thanks{peter@berkeley.edu; Computer Science Division \& Department of Statistics, UC Berkeley.}}
\begin{document}
\maketitle
\begin{abstract}
We present a generalization of the adversarial linear bandits framework, where the underlying losses are kernel functions (with an associated reproducing kernel Hilbert space) rather than linear functions. We study a version of the exponential weights algorithm and bound its regret in this setting. Under conditions on the eigendecay of the kernel we provide a sharp characterization of the regret for this algorithm. When we have polynomial eigendecay $\mu_j \le \mathcal{O}(j^{-\beta})$, we find that the regret is bounded by $\mathcal{R}_n \le \mathcal{O}(n^{\beta/(2(\beta-1))})$; while under the assumption of exponential eigendecay $\mu_j \le \mathcal{O}(e^{-\beta j })$, we get an even tighter bound on the regret $\mathcal{R}_n \le \mathcal{O}(n^{1/2}\log(n)^{1/2})$. We also study the full information setting when the underlying losses are kernel functions and present an adapted exponential weights algorithm and a conditional gradient descent algorithm.
\end{abstract}
\section{Introduction}
In adversarial online learning \citep{cesa2006prediction,hazan2016introduction}, a player interacts with an unknown and arbitrary adversary in a sequence of rounds. At each round, the player chooses an action from an action space and incurs a loss associated with that chosen action. The loss functions are determined by the adversary and are fixed at the beginning of each round. After choosing an action the player observes some feedback, which can help guide the choice of actions in subsequent rounds. The most common feedback model is the \emph{full information} model, where the player has access to the entire loss function at the end of each round. Another, more challenging feedback model is the \emph{partial information} or \emph{bandit} feedback model where the player at the end of the round just observes the loss associated with the action chosen in that particular round. There are also other feedback models in between and beyond the full and bandit information models, many of which have also been studied in detail. A figure of merit that is often used to judge online learning algorithms is the notion of \emph{regret}, which compares the players actions to the best single action in hindsight (defined formally in Section \ref{sec:notations}). 

When the underlying action space is a continuous and compact (possibly
convex) set and the losses are linear or convex functions over this
set; there are many algorithms known that attain sub-linear and
sometimes optimal regret in both these feedback settings. In this work
we present a generalization of the well studied adversarial online
linear learning framework. In our paper, at each round the player
selects an action $a \in \mathcal{A} \subset \mathbb{R}^d$. This
action is mapped to an element in a reproducing kernel Hilbert space
(RKHS) generated by a mapping $\ke (\cdot,\cdot)$. The function
$\ke(\cdot,\cdot)$ is a kernel map, that is, it can be thought of as the inner product of an appropriate Hilbert space $\mathcal{H}$. The kernel map can be expressed as $\ke(x,y) = \langle \Phi(x),\Phi(y) \rangle_\mathcal{H}$, where $\Phi(\cdot) \in \mathbb{R}^{D}$ is the associated feature map.

At each round the loss is $\langle \Phi(a), w \rangle_\mathcal{H}$, where $w \in \mathcal{H}$ is the adversary's action. In the full information setting, as feedback, the player has access to the entire adversarial loss function $\langle \cdot , w\rangle_{\mathcal{H}}$. In the bandit setting the player is only presented with the value of the loss, $\langle \Phi(a), w \rangle_\mathcal{H}$. 

Notice that this class of losses is much more general than ordinary linear losses and includes potentially non-linear and non-convex losses like:
\begin{enumerate}
    \item Linear Losses: $\langle a,w \rangle_\mathcal{H}= a^{\top}w$. This loss is well studied in both the bandit and full information settings. We shall see that our regret bounds will match the bounds established in the literature for these losses.
    \item Quadratic Losses: $\left\langle \Phi(a), \binom{W}{b} \right \rangle_{\mathcal{H}} = a^\top W a   + b^\top a$, where $W$ is a symmetric (not necessarily positive semi-definite) matrix and $b$ is a vector. Convex quadratic losses have been well studied under full information feedback as the online eigenvector decomposition problem. Our work establishes regret bounds in the full information setting and also in the previously unexplored bandit feedback setting.
    \item Gaussian Losses: $\langle \Phi(a), \Phi(y) \rangle_\mathcal{H} = \exp\left(-\lVert a - y \rVert_2^2/2\sigma^2 \right)$. We provide regret bounds for kernel losses not commonly studied before like Gaussian losses that provide a different loss profile than a linear or convex loss. 
    \item Polynomial Losses: $\langle \Phi(a), \Phi(y) \rangle_{\mathcal{H}} = (1+ x^{\top}y)^2$ for example. We also provide regret bounds for polynomial kernel losses which are potentially (non-convex) under both partial and full information settings. Specifically in the full information setting we study posynomial losses (discussion in Appendix \ref{sec:posynomials}).
\end{enumerate}

\subsection{Related Work}
Adversarial online convex bandits were introduced and first studied by \citet{kleinberg2005nearly,flaxman2005online}. The problem most closely related to our work is the case when the losses are linear and was introduced earlier by \citet{mcmahan2004online,awerbuch2004adaptive}. To improve the dimension dependence in the regret, \citet{dani2008price,cesa2012combinatorial,bubeck2012towards} proposed the \textsf{EXP 2} (Expanded Exp) algorithm under different choices of exploration distributions. \citet{dani2008price} worked with the uniform distribution over the barycentric spanner of the set, \citet{cesa2012combinatorial} used the uniform distribution over the set and \citet{bubeck2012towards} set the exploration distribution to be the one given by John's theorem which leads to a regret bound of $\mathcal{O}((dn\log(\lvert \mathcal{A} \rvert))^{1/2})$. Here $\lvert \mathcal{A}\rvert$ is the number of actions, $n$ is the number of rounds and $d$ is the dimension of the losses. In the case of linear bandits when the set $\mathcal{A}$ is convex and compact, \citet{abernethy2008competing} analyzed mirror descent to get a regret bound of $\mathcal{O}(d\sqrt{\theta n\log(n)})$ for some $\theta>0$. For the case with general convex losses with bandit feedback recently \citet{bubeck2016kernel} proposed a poly-time algorithm that has a regret guarantee of $\tilde{\mathcal{O}}(d^{9.5}\sqrt{n})$, which is optimal in its dependence on the number of rounds $n$. Previous work on this problem includes, \citet{agarwal2010optimal,saha2011improved,hazan2014bandit,dekel2015bandit,bubeck2015bandit,hazan2016optimal} in the adversarial setting under different assumptions on the structure of the convex losses and by \citet{agarwal2011stochastic} who studied this problem in the stochastic setting\footnote{For an extended bibliography of the work on online convex bandits see \citet{bubeck2016kernel}.}. \citet{valko2013finite} study stochastic kernelized contextual bandits with a modified Upper Confidence Bound (UCB) algorithm to obtain a regret bound similar to ours, $\mathcal{R}_n \le \tilde{d}\sqrt{n}$ where $\tilde{d}$ is the effective dimension dependent on the eigendecay of the kernel. This problem was also studied previously for loss functions drawn from Gaussian processes by \cite{srinivas2009gaussian}. Online learning under bandit feedback has also been studied when the losses are non-parametric, for example when the losses are Lipschitz \citep{rakhlin2015online,cesa2017algorithmic}. 

In the full information case, the online optimization framework with convex losses was first introduced by \citet{zinkevich2003online}. The conditional gradient descent algorithm (a modification of which we study in this work) for convex losses in this setting was introduced and analyzed by \citet{jaggi2011convex} and then improved subsequently by \citet{hazan2012projection}. The exponential weights algorithm which we modify and use multiple times in this paper has a rich history and has been applied to various online as well as offline settings. 
The case of convex quadratic losses has been well studied in the full information setting. This problem is called online eigenvector decomposition or online Principal Component Analysis (PCA). Recently \citet{allen2017follow} established a regret bound of $\tilde{\mathcal{O}}(\sqrt{n})$ for this problem by presenting an efficient algorithm that achieves this rate -- a modified exponential weights strategy termed ``follow the compressed leader''. Previous results for this problem were established in both adversarial and stochastic settings by modifications of exponential weights, gradient descent and follow the perturbed leader algorithms \citep{tsudamatrix,kalaiQuad,warmuth2006online,arora,warmuth2008randomized,garber2015online}. 

In the full information setting there has also been work on analyzing
gradient descent and mirror descent in an RKHS
\citep{mcmahan2014unconstrained,balandat2016minimizing}. However, in
these papers, the player is allowed to play any action in a bounded set in a Hilbert space, while in our paper the player is constrained to only play \emph{rank one} actions, that is the player chooses an action in $\mathcal{A}$ which gets mapped to an action in the RKHS.


\subsubsection*{Contributions}
Our primary contribution is to extend the linear bandits framework to more general classes of kernel losses. We present an exponential weights algorithm in this setting and establish a regret bound on its performance. We provide a more detailed analysis of the regret under assumptions on the eigendecay of the kernel. When we assume polynomial eigendecay of the kernel ($\mu_j \le \mathcal{O}(j^{-\beta})$) we can guarantee the regret is bounded as $\mathcal{R}_n\le \mathcal{O}\left(n^{\frac{\beta}{2(\beta-1)} }\right)$. Under exponential eigendecay we can guarantee an even sharper bound on the regret of $\mathcal{R}_n \le \mathcal{O}(n^{1/2}\log(n)^{1/2})$. We also provide an exponential weights algorithm and a conditional gradient algorithm for the full information case where we don't need to assume any conditions on the eigendecay. Finally we provide a couple of applications of our framework -- (i) general quadratic losses (not necessarily convex) with linear terms which we can solve efficiently both in the full information setting and the bandit setting, (ii) we provide a computationally efficient algorithm when the underlying losses are posynomials (special class of polynomials).
\subsubsection*{Organization of the Paper}
In the next section we introduce the notation and definitions. In Section \ref{sec:banditintro} we present our algorithm under bandit feedback along with regret bounds for it. In Section \ref{sec:fullinfo} we study the problem in the full information setting. In Section \ref{sec:applications} we apply our framework to general quadratic losses prove that our algorithms are computationally efficient in this setting. All the proofs and technical details are relegated to the appendix. Also in the appendix is the example of our framework applied to posynomial losses and experimental evidence verifying our claims.

\subsection{Notation, main definitions and setting}
\label{sec:notations}
Here we introduce definitions and notational conventions used throughout the paper.

In each round $t= \{1,\ldots,n\}$, the player chooses an action vector $\{a_t\}_{t=1}^{n} \in \mathcal{A} \subset \mathbb{R}^d$. The underlying kernel function at each round is  $\ke(\cdot,\cdot)$ which is a map from $\mathbb{R}^d \times \mathbb{R}^d \to \mathbb{R}$ such that it is a \emph{kernel map} and has an associated separable reproducing kernel Hilbert space (RKHS) $\mathcal{H}$ with an inner product $\langle \cdot, \cdot \rangle_\mathcal{H}$ \citep[for more properties of kernel maps and RKHS see][]{scholkopf2001learning}. Let $\Phi(\cdot): \mathbb{R}^d \mapsto \mathbb{R}^{D}$ denote a \emph{feature map} of $\ke(\cdot,\cdot)$ such that for every $x,y$ we have $\ke(x,y) = \langle \Phi(x),\Phi(y) \rangle_\mathcal{H}$. Note that the dimension of the RKHS, $D$ could be infinite (for example in the Gaussian kernel over $\left[0,1\right]^{d}$). 

We let the adversary choose a vector in $\mathcal{H}$, $w_t \in \mathcal{W} \subset \mathbb{R}^D$ and at each round the loss incurred by the player is $\langle a_t,w_t\rangle_\mathcal{H}$. We assume that the adversary is oblivious, that is, it is a function of the previous actions of the player $(a_1,\ldots,a_{t-1})$ but unaware of the randomness used to generate $a_t$. 
We let the size of the sets $\mathcal{A},\mathcal{W}$ be bounded\footnote{We set the bound on the size of both sets to be the same for ease of exposition, but they could be different and would only change the constants in our results.} in \emph{kernel norm}, that is, 
\begin{align}\label{eq:setbound}
\sup_{a \in  \mathcal{A}} \ke(a,a) \le \mathcal{G}^2 \text{\qquad and, } \sup_{w \in \mathcal{W}} \langle w,w\rangle_\mathcal{H} \le \mathcal{G}^2.
\end{align}
Throughout this paper we assume a \emph{rank-one learner}, that is, in each round the player can pick a vector $v \in \mathcal{H}$, such that $v = \Phi(a)$ for some $a \in \mathbb{R}^d$. We now formally define the notion of expected regret.
\begin{definition}[Expected regret] \label{def:regret}
The expected regret of an algorithm $\mathcal{M}$ after $n$ rounds is defined as
\begin{align}
\label{eq:regretdefinition}
\mathcal{R}_n & = \mathbb{E}_{\mathcal{M}} \left[\sum_{t=1}^{n} \langle \Phi(a_t),w_t \rangle_\mathcal{H} -\sum_{t=1}^{n} \langle \Phi(a^*),w_t \rangle_\mathcal{H} \right],
\end{align}
where $a^* = \inf_{a\in\mathcal{A}}\left\{ \sum_{t=1}^{n} \langle \Phi(a),w_t\rangle_\mathcal{H} \right\}$ and the expectation is over the randomness in the algorithm. 
\end{definition}
Essentially this amounts to comparing against the \emph{best single action} $a^*$ in hindsight. Our hope will be to find a randomized strategy such that the regret grows sub-linearly with the number of rounds $n$. In what follows we will omit the subscript $\mathcal{H}$ from the subscript of the inner product whenever it is clear from the context that it refers to the RKHS inner product.

To establish regret guarantees we will find that it is essential to work with finite dimensional kernels when working under bandit feedback (more details regarding this are in the proof of the regret bound of Algorithm \ref{a:expweightskernels}). General kernel maps can have infinite dimensional feature maps thus we will require the construction of a finite dimensional kernel that uniformly approximates the original kernel $\ke(\cdot,\cdot)$. This motivates the definition of $\epsilon$-approximate kernels.
\begin{definition}[$\epsilon$-approximate kernels] \label{def:epsilonapproximate}
Let $\ke_1$ and $\ke_2$ be two kernels over $\mathcal{A} \times \mathcal{A}$ and let $\epsilon >0$. We say $\ke_2$ is an $\epsilon$-approximation of $\ke_1$ if for all $x, y \in \mathcal{A}$, $\lvert \ke_1(x,y) - \ke_2(x,y)\rvert \leq \epsilon.$
\end{definition}

\section{Bandit Feedback Setting}
\label{sec:banditintro}
In this section we present our results on \emph{kernel bandits}. In
the bandit setting we assume the player knows the underlying kernel
function $\ke(\cdot,\cdot)$, however, at each round after the player
plays a vector $a_t$ only the value of the loss associated with that
action is revealed to the player -- $\langle
\Phi(a_t),w_t\rangle_{\mathcal{H}}$ -- and not the action of the
adversary $w_t$. We also assume that the player's action set
$\mathcal{A}$ has finite cardinality\footnote{This assumption can be
relaxed to let $\mathcal{A}$ be a compact set when $\ke$ is Lipschitz
continuous. In this setting we can instead work with an appropriately
fine approximating cover of the set $\mathcal{A}$.}.This is a
generalization of the well studied adversarial linear bandits problem.
As we will see in subsequent sections, to guarantee a bound on the regret in the bandit setting our algorithm will build an estimate of adversary's action $w_t$. This becomes impossible if $w_t$ is infinite dimensional ($D \to \infty$). To circumvent this, we will construct a finite dimensional proxy kernel that is an $\epsilon$-approximation of $\ke$.

Whenever no approximate kernel is needed, for example when $D < \infty$ we allow the adversary to be able to choose an action $w_t \in \mathcal{W} \subset \mathbb{R}^D$ without imposing extra requirements on the set $\mathcal{W}$ other than it being bounded in $\mathcal{H}$ norm. When $D$ is infinite we impose an additional constraint on the adversary to also select \emph{rank-one} actions at each round, that is, $w_t = \Phi(y_t)$ for some $y_t \in \mathbb{R}^d$. Next we present a procedure to construct a finite kernel that approximates the original kernel well.

\subsection{Construction of the finite dimensional kernel}
\label{sec:constrproxy} 
To construct the finite dimensional kernel we will rely crucially on Mercer's theorem. We first recall a couple of useful definitions.
\begin{definition}\label{def:L2class}
Let $\mathcal{A} \subset \mathbb{R}^d$ and $\mathbb{P}$ be a probability measure supported over $\mathcal{A}$. Let $L_2(\mathcal{A};\mathbb{P})$ denote square integrable functions over $\mathcal{A}$ and measure $\mathbb{P}$, $L_2(\mathcal{A};\mathbb{P}) := \left\{f:\mathcal{A} \rightarrow \mathbb{R} \Bigg \lvert \int_{\mathcal{A}} f^2(x) d\mathbb{P}(x) < \infty \right\}$.
\end{definition}

\begin{definition}\label{def:squareinte}
A kernel $\ke: \mathcal{A} \times \mathcal{A} \to \mathbb{R}$ is square integrable with respect to measure $\mathbb{P}$ over $\mathcal{A}$, if $\int_{\mathcal{A} \times \mathcal{A}} \ke^2(x,y) d\mathbb{P}(x) d\mathbb{P}(y) < \infty $.
\end{definition}
Now we present Mercer's theorem \citep{mercer1909functions} \citep[see][]{cristianini2000introduction}.
\begin{theorem}[Mercer's Theorem]\label{thm:mercertheo}
Let $\mathcal{A} \subset \mathbb{R}^d$ be compact and $\mathbb{P}$ be a finite Borel measure with support $\mathcal{A}$. Suppose $\ke$ is a continuous square integrable positive definite kernel on $\mathcal{A}$, and define a positive definite operator $\mathcal{T}_{\ke} : L_2(\mathcal{A};\mathbb{P}) \mapsto L_2(\mathcal{A};\mathbb{P})$ by
\begin{align*}
\left( \mathcal{T}_\ke f    \right) \left( \cdot \right) := \int_{\mathcal{A}} \ke(\cdot, x) f(x) \diff \mathbb{P}.
\end{align*}
Then there exists a sequence of eigenfunctions $\{ \phi_i\}_{i=1}^\infty$ that form an orthonormal basis of $L_2(\mathcal{A};\mathbb{P})$ consisting of eigenfunctions of $\mathcal{T}_{\ke}$, and an associated sequence of non-negative eigenvalues $\{\mu_j\}_{j=1}^\infty$ such that $\mathcal{T}_{\mathcal{K}}(\phi_j) = \mu_j \phi_j$ for $j = 1,2,\ldots$. Moreover the kernel function can be represented as
\begin{equation}\label{eq:Mercerthm}
\ke(u,v) = \sum_{i=1}^\infty  \mu_i \phi_i(u)\phi_i(v),
\end{equation}
where the convergence of the series holds uniformly.
\end{theorem}
Mercer's theorem suggests a natural way to construct a feature map $\Phi(x)$ for $\ke$ by defining the $i^{th}$ component of the feature map to be $\Phi(x)_i := \sqrt{\mu_i} \phi_i(x)$. Under this choice of the feature map the eigenfunctions $\{\phi_i\}_{i=1}^{\infty}$ are orthogonal with respect to the inner product $\langle\cdot ,\cdot \rangle_{\mathcal{H}}$\footnote{To see this observe that the function $\phi_i$ can be expressed as a vector in the RKHS as a vector $v_i$ with $\phi_i$ in the $i^{th}$ coordinate and zeros everywhere else. So for any two $v_i$ and $v_j$ with $i\neq j$ we have $\langle v_i,v_j \rangle_{\mathcal{H}} = 0$.}. Armed with Mercer's theorem we first present a simple deterministic procedure to obtain a finite dimensional $\epsilon$-approximate kernel of $\ke$. When the eigenfunctions of the kernel are uniformly bounded, $\sup_{x \in \mathcal{A}}\lvert \phi_j(x) \rvert \le \mathcal{B}$ for all $j$, and if the eigenvalues decay at a suitable rate we can truncate the series in Equation \eqref{eq:Mercerthm} to get a finite dimensional approximation.
\begin{restatable}{lemma}{detelemma}
\label{lemma_chopped_kernel} Given $\epsilon >0$, let $\{ \mu_j \}_{j=1}^\infty$ be the Mercer operator eigenvalues of $\ke$ under a finite Borel measure $\mathbb{P}$ with support $\mathcal{A}$ and eigenfunctions $\{ \phi_j \}_{j=1}^\infty$ with $\mu_1 \geq \mu_2 \geq \cdots$. Further assume that $\sup_{j \in \mathbb{N}} \sup_{x \in \mathcal{A}} \lvert \phi_j(x) \rvert \leq \mathcal{B}$ for some $\mathcal{B} < \infty$. Let $m(\epsilon)$ be such that $\sum_{j=m+1}^\infty \mu_j \leq \frac{\epsilon}{4\mathcal{B}^2}$. Then the truncated feature map,
\begin{equation}
\Phi^o_m(x) := \begin{cases}  \sqrt{\mu_i}\phi_i(x) & \text{if } i \leq m \\ 0 & \text{o.w.} \end{cases}
\end{equation}
induces a kernel $\kc^o(x,y) := \langle \Phi^{o}_m(x),\Phi^{o}_m(y) \rangle_{\mathcal{H}} = \sum_{j=1}^m \mu_j \phi_j(x)\phi_j(y)$, for all  $(x,y) \in \mathcal{A}\times \mathcal{A}$ that is an $\epsilon/4$-approximation of $\ke$.
\end{restatable}
The Hilbert space induced by $\kc^{o}$ is a subspace of the original Hilbert space $\mathcal{H}$. The proof of this lemma is a simple application of Mercer's theorem and is relegated to Appendix \ref{app:proxykernel}. If we have access to the eigenfunctions of $\ke$ we can construct and work with $\kc^o$ because as Lemma \ref{lemma_chopped_kernel} shows $\kc^o$ is an $\epsilon/4$-approximation to $\ke$. Additionally, $\kc^o$ also has the same first $m$ Mercer eigenvalues and eigenfunctions under $\mathbb{P}$ as $\ke$. Unfortunately, in most applications of interest the analytical computation of the eigenfunctions $\{ \phi_i \}_{i=1}^{\infty}$ is not possible. We can get around this by building an estimate of the eigenfunctions using samples from $\mathbb{P}$ by leveraging results from kernel principal component analysis (PCA). 
\begin{definition}Let $S_m$ be the subspace of $\mathcal{H}$ spanned by the first $m$ eigenvectors of the covariance matrix $\mathbb{E}_{x \sim \mathbb{P}}\left[ \Phi(x) \Phi(x)^\top  \right]$.
\end{definition}
This corresponds\footnote{This holds as the $i^{th}$ eigenvector of the covariance matrix has $\phi_i$ as the $i^{th}$ coordinate and zero everywhere else combined with the fact that $\{\phi_i \}_{i=1}^{\infty}$ are orthonormal under the $L_2(\mathcal{A};\mathbb{P})$ inner product.}
to the span of the eigenfunctions $\phi_1,\ldots,\phi_m$ in $\mathcal{H}$. Define the linear projection operator $P_{S_m}: \mathcal{H} \mapsto \mathcal{H}$ that projects onto the subspace $S_m$; where $P(S_m)(v+ v^{\perp}) = v$, if $v \in S_m$ and $v^{\perp} \in S_m^{\perp}$. 
\begin{remark}
The feature map $\Phi^o_m(x)$ is a projection of the complete feature map to this subspace, $\Phi^o_m(x) = P_{S_m}(\Phi(x))$.
\end{remark}

Let $x_1,x_2,\ldots,x_p \sim \mathbb{P}$ be $p$ i.i.d. samples and construct the sample (kernel) covariance matrix, $\hat{\Sigma} := \frac{1}{p}\sum_{i=1}^p \Phi(x_i) \Phi(x_i)^{\top}$. Let $\hat{S}_m$ be the subspace spanned by the $m$ top eigenvectors of $\hat{\Sigma}$. Define the stochastic feature map, $\Phi_m(x) := P_{\hat{S}_m}(\Phi(x))$, the feature map defined by projecting $\Phi(x)$ to the random subspace $\hat{S}_m$. Intuitively we would expect that if the number of samples $p$ is large enough, then the  kernel defined by the feature map $\Phi_m(x)$, $\kc(x,y) = \langle \Phi_m(x), \Phi_m(y) \rangle_{\mathcal{H}}$ will also be an $\epsilon$-approximation for the original kernel $\ke$. Formalizing this claim is the following theorem.

\begin{theorem}\label{thm:proxy_properties1}
Let $\epsilon, m, \mathbb{P}$ be defined as in Lemma \ref{lemma_chopped_kernel} and let the $m$-th level eigen-gap be $\delta_m := \frac{1}{2}\left( \mu_m - \mu_{m+1}  \right)$. Further let $\alpha>0$, $B_m = \frac{2\mathcal{G}^2}{\delta_m} \left( 1 + \sqrt{ \frac{\alpha}{2}}   \right)$ and $p \geq \frac{4B_m^2 \mathcal{G}^2 }{(\min(\sqrt{\epsilon},\delta_m))^2} $. Then the finite dimensional kernels $\hat{\ke}_m^o$ and $\hat{\ke}_m$ satisfy the following properties with probability $1-e^{-\alpha}$,
\begin{enumerate}
\item $\sup_{x,y \in \mathcal{A}} |\ke(x,y) - \hat{\ke}_m(x,y) | \leq \epsilon$.
\item The Mercer eigenvalues $\mu^{(p)}_1 \geq \cdots \geq \mu^{(p)}_m$ and $\mu_1 \geq \cdots \geq \mu_m$ of $\kc$ and $\kc^o$ are close, $\sup_{i=1, \cdots, m} |\mu^{(p)}_i - \mu_i| \leq \epsilon/4$.
\end{enumerate}
\end{theorem}
Theorem \ref{thm:proxy_properties1} shows that given $\epsilon> 0$ the finite dimensional proxy $\kc$ is a $\epsilon$-approximation of $\ke$ with high probability as long as sufficiently large number of samples are used.  Furthermore, the top $m$ eigenvalues of the second moment matrix of $\ke$ are at most $\sqrt{\epsilon}/2$-away from the eigenvalues of the second moment matrix of $\kc$ under $\mathbb{P}$.

\begin{function}[t]
\caption{ProxyKernel($\mathcal{K},m,\mathcal{A},\mathbb{P},p$)} \label{a:algoproxy}
	\KwIn{Kernel $\ke$, effective dimension $m$, set $\mathcal{A}$, measure $\mathbb{P}$, number of samples $p$.}
	\SetKwInOut{Output}{Function}
	\KwOut{Finite proxy feature map $\Phi_m(\cdot)$}
			Sample $x_1, \cdots, x_p \sim \mathbb{P}$. 

Construct sample Gram matrix $\hat{\mathbb{K}}_{i,j} = \frac{1}{p} \ke(x_i,x_j)$.

Calculate the top $m$ eigenvectors of $\hat{\mathbb{K}} \to \{\omega_1,\omega_2,\ldots,\omega_m \}$.

\For {$j=1,\ldots,m$}  
      {
      Set $v_j \leftarrow \sum_{k=1}^{p} \omega_{jk}\Phi(x_k)/(\lv \sum_{k=1}^{p} \omega_{jk}\Phi(x_k)\rv_2)$   \qquad \qquad ($\omega_{jk}$ is the $k^{th}$ component of $\omega_j$)
      }

Define the feature map 
\begin{align*}
\Phi_m(\cdot) := \begin{bmatrix}
\langle v_1,\Phi(\cdot) \rangle_{\mathcal{H}} \\
\vdots \\
\langle v_m,\Phi(\cdot) \rangle_{\mathcal{H}} 
\end{bmatrix} = \begin{bmatrix}
\sum_{k=1}^p \omega_{1k}\ke(x_k,\cdot)/(\lv \sum_{k=1}^{p} \omega_{1k}\Phi(x_k)\rv_2) \\
\vdots \\
\sum_{k=1}^p \omega_{mk}\ke(x_k,\cdot)/(\lv \sum_{k=1}^{p} \omega_{mk}\Phi(x_k)\rv_2)
\end{bmatrix}.
\end{align*}
	
\end{function}
%
%
%
%

To construct $\Phi_m(\cdot)$ we need to calculate the top $m$ eigenvectors of the sample covariance matrix $\hat{\Sigma}$, however, it is equivalent to calculate the top $m$ eigenvectors of the sample Gram matrix $\mathbb{K}$ and use them to construct the eigenvectors of $\hat{\Sigma}$ (for more details see Appendix \ref{app:kernelPCA} where we review the basics of kernel PCA).

\label{s:resultsmain}
\subsection{Bandits Exponential Weights}
\label{s:algobandits}
In this section we present a modified version of exponential weights adapted to work with kernel losses. Exponential weights has been analyzed extensively applied to linear losses under bandit feedback \citep{dani2008price,cesa2012combinatorial,bubeck2012towards}. Two technical challenges make it difficult to directly adapt their algorithms to our setting.

The first challenge is that at each round we need to estimate the adversarial action $w_t$. If the feature map of the kernel is finite dimensional this is easy to handle, however when the feature map is infinite dimensional, this becomes challenging and we need to build an approximate feature map $\Phi_m(\cdot)$ using Function \ref{a:algoproxy}. This introduces bias in our estimate of the adversarial action $w_t$ and we will need to control the contribution of the bias in our regret analysis. The second challenge will be to lower bound the minimum eigenvalue of the kernel covariance matrix as we will need to invert this matrix to estimate $w_t$. For general kernels which are infinite dimensional, the minimum eigenvalue is zero. To resolve this we will again turn to our construction of a finite dimensional proxy kernel. 

\subsection{Bandit Algorithm and Regret Bound} \label{ss:banditregretalgo}
\begin{algorithm}[t]
\SetKwInOut{Input}{Input}
    \SetKwInOut{Output}{Output}
        \Input{Set $\mathcal{A}$, learning rate $\eta > 0$, mixing coefficient $\gamma>0$, number of rounds $n$, uniform distribution $\nu_{unif}$ over $\mathcal{A}$, exploration distribution $\nu_{\mathcal{J}}^{\mathcal{A}}$ over $\mathcal{A}$, kernel map $\ke$, effective dimension $m(\epsilon)$, number of samples $p$.}
    Build kernel $\kc$ with feature map $\Phi_m(\cdot) =$ \ref{a:algoproxy}$(\ke,m,\mathcal{A},\nu_{unif},p)$. \\
    Set $q_1(a) = \nu_{\mathcal{J}}^{\mathcal{A}}$.
    
    \For {$t=1,\ldots,n$}  
      {
      	Set $p_t = \gamma \nu_{\mathcal{J}}^{\mathcal{A}} + (1-\gamma) q_t$.
      
        Choose $a_t \sim p_t$.
      
        Observe $\langle \Phi(a_t),w_t \rangle_{\mathcal{H}}$.
        
        Build the covariance matrix
        \begin{align*}
        \Sigma_m^{(t)} = \mathbb{E}_{x \sim p_t}\left[\Phi_m(x)\Phi_m(x)^\top \right].
        \end{align*}
        
        Compute the estimate $\hat{w}_t = \left(\Sigma_m^{(t)}\right)^{-1}\Phi_m(a_t) \langle \Phi(a_t), w_t \rangle_{\mathcal{H}}$.
        
        Update $q_{t+1}(a) \propto q_t(a) \cdot\exp\left(-\eta \cdot \langle \hat{w}_t, \Phi_m(a) \rangle_{\mathcal{H}} \right)$.
        }
   \caption{Bandit Information: Exponential Weights \label{a:expweightskernels}}
  \end{algorithm} 
In our exponential weights algorithm we first build the finite dimensional proxy kernel $\kc$ using Function \ref{a:algoproxy}. The rest of the algorithm is then almost identical to the exponential weights algorithm (EXP 2) studied for linear bandits. In Algorithm \ref{a:expweightskernels} we set the exploration distribution $\nu_{\mathcal{J}}^{\mathcal{A}}$ to be such that it induces John's distribution ($\nu_{\mathcal{J}}$) over $\Phi_m(\mathcal{A}) : = \{\Phi_m(a) \in \mathbb{R}^{m} : a \in \mathcal{A}\}$ (first introduced as an exploration distribution by \citet{bubeck2012towards}; a short discussion is presented in Appendix \ref{app:Johntheorem}). 
Note that for finite sets it is possible to build an approximation to minimal volume ellipsoid containing $conv(\Phi_m(\mathcal{A}))$--John's ellipsoid and John's distribution in polynomial time \citep{grotschel2012geometric}\footnote{It is thus possible to construct $\nu_{\mathcal{J}}$ over $\Phi_m(\mathcal{A})$ in polynomial time. However, as $\mathcal{A}$ is a finite set, using $\Phi_m(\cdot)$ and $\nu_{\mathcal{J}}$ it is also possible to construct $\nu_{\mathcal{J}}^{\mathcal{A}}$ over $\mathcal{A}$ efficiently.}. In this section we assume that the set $\mathcal{A}$ is such that the John's ellipsoid is centered at the origin.

Crucially note that we construct the finite dimensional feature map $\Phi_m(\cdot)$ only \emph{once} before the first round. In the algorithm presented above we build $\Phi_m(\cdot)$ using the uniform distribution $\nu_{unif}$ over $\mathcal{A}$ assuming that kernel $\ke$ has fast eigendecay under this measure. However, any other distribution say -- $\nu_{alt}$ (with support $\mathcal{A}$) could also be used instead of $\nu_{unif}$ in Algorithm \ref{a:expweightskernels} if the kernel $\ke$ enjoys fast eigendecay under $\nu_{alt}$. 

In our algorithm we build and invert the exact covariance matrix $\Sigma_m^{(t)}$, however this can be relaxed and we can instead work with a sample covariance matrix. We analyze the required sample complexity and error introduced by this additional step in Appendix \ref{app:covariancesample}. We now state the main result of this paper which is an upper bound on the regret of Algorithm \ref{a:expweightskernels}.

\begin{theorem}\label{thm:mainregretbound} Let $\mu_i$ be the $i$-th Mercer operator eigenvalue of $\ke$ for the uniform measure $\nu_{unif}$ over $\mathcal{A}$. Let $\epsilon \le \mathcal{G}^2$ and let $m,p,\alpha$ be chosen as specified by the conditions in Theorem \ref{thm:proxy_properties1}. Let the mixing coefficient, $\gamma = 4\eta\mathcal{G}^4m$. Then Algorithm \ref{a:expweightskernels} with probability $1 - e^{-\alpha}$ has regret bounded by
\begin{align*}
\mathcal{R}_n \le 4\gamma \mathcal{G}^2 n + (e-2)\mathcal{G}^4\eta m n + 2\epsilon n + 
\frac{2\epsilon n}{\mathcal{G}^2\eta}+ \frac{1}{\eta}\log(\lvert \mathcal{A} \rvert).
\end{align*}
\end{theorem}
We prove this theorem in Appendix \ref{app:regretboundbandit}. Note that this is similar to the regret rate attained for adversarial linear bandits \citep{dani2008price,cesa2012combinatorial,bubeck2012towards} with additional terms that accounts for the bias in our loss estimates $\hat{w}_t$. In our regret bounds the parameter $m$ plays the role of the effective dimension and will be determined by the rate of the eigendecay of the kernel. When the underlying Hilbert space is finite dimensional (as is the case when the losses are linear or quadratic) our regret bound recovers exactly the results of previous work (that is, $\epsilon = 0$ and $m=d$). 

We note that the exploration distribution can also be chosen to be the uniform distribution over the Barycentric spanner of $\Phi_m(\mathcal{A})$. But this choice leads to slightly worse bounds on the regret and we omit a detailed discussion here for the sake of clarity. Next we state the following different characteristic eigenvalue decay profiles.
\begin{definition}[Eigenvalue decay] Let the Mercer operator eigenvalues of a kernel $\ke$ with respect to a measure $\mathbb{P}$ over a set $\mathcal{A}$ be denoted by $\mu_1 \ge \mu_2 \ge \ldots$.
\begin{enumerate}
\item $\ke$ is said to have $(C,\beta)$-\textbf{polynomial eigenvalue decay} (with $C>0, \beta>1$) if for all $j \in \mathbb{N}$ we have $\mu_j  \le  Cj^{-\beta}$.
\item $\ke$ is said to have $(C,\beta)$-\textbf{exponential eigenvalue decay} (with $C>0, \beta>0$) if for all $j \in \mathbb{N}$ we have $\mu_j  \le  Ce^{-\beta j}$.
\end{enumerate}
\end{definition}
Under assumptions on the eigendecay we can establish bounds on the \emph{effective dimension} $m$ and $\mu_m$, so that the condition stated in Lemma \ref{lemma_chopped_kernel} is satisfied and we are guaranteed to build an $\epsilon$-approximate kernel $\kc$. We establish bounds on $m$ in Proposition \ref{prop:mbounds} presented in Appendix \ref{app:effectivedim}. Under the eigendecay profiles stated above we can now invoke Theorem \ref{thm:mainregretbound}. 
\begin{corollary}\label{cor:specificregretbound} Let the conditions stated in Theorem \ref{thm:mainregretbound} hold and let $\mathcal{G}=1$. Then Algorithm \ref{a:expweightskernels} has its regret bounded by the following rates with probability $1-e^{-\alpha}$. 
\begin{enumerate}
\item If $\ke$ has $(C,\beta)$-polynomial eigenvalue decay under the uniform measure $\nu_{unif}$, with $\beta >2$. Then by choosing the step size $\eta = \sqrt{\epsilon/(10m)}$ where, $\epsilon = \log(\lvert \mathcal{A} \rvert )/(2n)$ and $m = \left [\frac{4C\mathcal{B}^2}{(\beta-1)\epsilon}\right]^{1/\beta-1}$, with $n$ large enough such that $\eta<1$ and $\epsilon<1$, the expected regret is bounded by
\begin{align*}
\mathcal{R}_n & \le  \sqrt{160}\cdot \left[\frac{2^{\beta+2}C\mathcal{B}^2}{\beta-1}\right]^{\frac{1}{2(\beta-1)}}\cdot \left(\log(\lvert A \rvert)\right)^{\frac{\beta-2}{2(\beta-1)}}\cdot n^{\frac{\beta}{2(\beta-1)}}.
\end{align*}
\item If $\ke$ has $(C,\beta)$-exponential eigenvalue decay under the uniform measure $\nu_{unif}$. Then by choosing the step-size $\eta = \sqrt{\epsilon/(10m)}$ where, $\epsilon = \log(\lvert \mathcal{A} \rvert )/(2n)$ and $m = \frac{1}{\beta}\log\left(\frac{4C \mathcal{B}^2}{\beta\epsilon}\right)$ with $n$ large enough so that $\epsilon<1$, the expected regret is bounded by
\begin{align*}
\mathcal{R}_n & \le \sqrt{\frac{320\cdot\log(\lvert \mathcal{A} \rvert)}{\beta}\cdot \log\left( \frac{40C\mathcal{B}^2 n }{\beta \log(\lvert \mathcal{A} \rvert)}\right)\cdot  n}.
\end{align*}
\end{enumerate}
\end{corollary}
\begin{remark} Under $(C,\beta)$-polynomial eigendecay condition we have that the regret is upper bounded by $\mathcal{R}_n \le \mathcal{O}(n^{\frac{\beta}{2(\beta-1)}})$. While when we have $(C,\beta)$-exponential eigendecay we almost recover the adversarial linear bandits regret rate (up to logarithmic factors), with $\mathcal{R}_n \le \mathcal{O}(n^{1/2}\log(n))$.
\end{remark}
In the corollary above we assume that $\mathcal{G} = 1$ for ease of exposition; results follow in a similar vein for other values of $\mathcal{G}$ with the constants altered. One way to interpret the results of Corollary \ref{cor:specificregretbound} in contrast to the regret bounds obtained for linear losses is the following. We introduce additional parameters into our analysis to handle the infinite dimensionality of our feature vectors -- the effective dimension $m$ and bias of our estimate $\epsilon$. When the effective dimension $m$ is chosen to be large we can build an estimate of the adversarial action $\hat{w}_t$ which has low bias, however this estimate would have large variance ($\mathcal{O}(m)$). On the other hand if we choose $m$ to be small we can build a low variance estimate of the adversarial action but with high bias ($\epsilon$ is large). We trade these off optimally to get the regret bounds established above. In the case of exponential decay we obtain that the choice $m = \mathcal{O}(\log(n))$ is optimal, hence the regret bound degrades only by a logarithmic factor in terms of $n$ as compared to linear losses (where $m$ would be a constant). When we have polynomial decay, the effective dimension is higher $m = \mathcal{O}(n^{\frac{1}{\beta-1}})$ which leads to worse bounds on the expected regret. Note that asymptotically as $\beta \to \infty$ the regret bound goes to $n^{1/2}$ which aligns well with the intuition that the effective dimension is small. While when $\beta \to 2$ (the effective dimension $m$ is large) and the regret upper bound becomes close to linear in $n$.

\section{Full Information Setting} \label{sec:fullinfo}
\subsection{Full information Exponential Weights}
\label{s:algosec}
\begin{algorithm}[h] 
\SetKwInOut{Input}{Input}
    \SetKwInOut{Output}{Output}
        \Input{Set $\mathcal{A}$, learning rate $\eta > 0$, number of rounds $n$.}
    Set $p_1(a)$ to be the uniform distribution over $\mathcal{A}$.
    
    \For {$t=1,\ldots,n$} 
      {
      
        Choose $a_t \sim p_t$
      
        Observe $\langle \Phi(a_t),w_t \rangle_{\mathcal{H}}$
        
        Update $p_{t+1}(a) \propto p_t(a) \cdot\exp\left(-\eta \cdot \langle w_t, \Phi(a) \rangle_{\mathcal{H}} \right)$
        }
        \caption{Full Information: Exponential Weights \label{a:expweightskernelsfull}}
\end{algorithm} 
We begin by presenting a version of the exponential weights algorithm, Algorithm \ref{a:expweightskernelsfull} adapted to our setup. In each round we sample an action vector $a_t \in \mathcal{A}$ (a compact set) from the exponential weights distribution $p_t$. After observing the loss, $\langle \Phi(a_t),w_t \rangle_{\mathcal{H}}$ we update the distribution by a multiplicative factor, $\exp(-\eta \langle w_t, \Phi(a) \rangle_{\mathcal{H}})$. In the algorithm presented we choose the initial distribution $p_1(a)$ to be uniform over the set $\mathcal{A}$, however we note that alternate initial distributions with support over the whole set could also be considered. We can establish a sub-linear regret of $\mathcal{O}(\sqrt{n})$ for the exponential weights algorithm.
\begin{theorem}
\label{t:expweightsregret}[Proof in Appendix \ref{app:regretexp}] Assume that in Algorithm \ref{a:expweightskernelsfull} the step size $\eta$ is chosen to be, $\eta  = \sqrt{\frac{\log(vol(\mathcal{A}))}{e-2}}\cdot \frac{1}{\mathcal{G}^2n^{1/2}}$,
with $n$ large enough such that $\sqrt{\frac{\log(vol(\mathcal{A}))}{e-2}}\frac{1}{n^{1/2}} \le 1$. Then the expected regret after $n$ rounds is bounded by, $$\mathcal{R}_n  \le \sqrt{(e-2)\log(vol(\mathcal{A}))}\mathcal{G}^2n^{1/2}.$$
\end{theorem}
\subsection{Conditional Gradient Descent}
\begin{algorithm}[t]
\SetKwInOut{Input}{Input}
    \SetKwInOut{Output}{Output}
    \Input{Set $\mathcal{A}$, number of rounds $n$, initial action $a_{1} \in \mathcal{A}$, inner product $\langle \cdot,\cdot \rangle_{\mathcal{H}}$, learning rate $\eta $, mixing rates $\{\gamma_{t} \}_{t=1}^n$. }
    $X_1 = \Phi(a_1)$
    
    choose $\mathcal{D}_1$ such that $\mathbb{E}_{x\sim \mathcal{D}_1}\Phi(x) = X_1$
    
    \For {$t=1,2,\ldots, n$} 
      {
         sample $a_t \sim \mathcal{D}_t$ 
         
         observe the loss  $\langle \Phi(a_t),w_t \rangle_{\mathcal{H}}$
         
         define $F_{t}(Y) \triangleq \eta \sum_{s=1}^{t-1} \langle  w_s , Y \rangle_{\mathcal{H}} + \lVert Y - X_1 \rVert_{\mathcal{H}}^{2}$
         
         compute $v_t = \argmin_{a \in \mathcal{A}} \langle \nabla F_t(X_t), \Phi(a) \rangle_{\mathcal{H}} $
         
         update mean $X_{t+1} = (1-\gamma_{t})X_{t} + \gamma_t \Phi(v_t)$
         
         choose $\mathcal{D}_{t+1}$ s.t. $\mathbb{E}_{x \sim \mathcal{D}_{t+1}}[\Phi(x)] = X_{t+1}$.
        }        
\caption{Full Information: Conditional Gradient Method \label{a:CG_algo}}  
\end{algorithm}
Next we present an online conditional gradient (Frank-Wolfe) method \citep{hazan2012projection} adapted for kernel losses. The conditional gradient method is also a well studied algorithm studied in both the online and offline setting \citep[for a review see][]{hazan2016introduction}. The main advantage of the conditional gradient method is that as opposed to projected gradient descent and related methods, the projection step is avoided. At each round the conditional gradient method involves the optimization of a linear (kernel) objective function over $\mathcal{A}$ to get a point $v_t \in \mathcal{A}$. Next we update the \emph{optimal mean} action $X_{t+1}$ by re-weighting the previous mean action $X_t$ by $(1-\gamma_t)$ and weight our new action $v_t$ by $\gamma_t$. Note that this construction also automatically suggests a distribution over $a_1,v_1,v_2,\ldots,v_t \in \mathcal{A}$ such that, $X_{t+1}$ is a convex combination of $\Phi(a_1),\Phi(x_1),\ldots,\Phi(a_t)$. For this algorithm we can prove a regret bound of $\mathcal{O}(n^{3/4})$.
\begin{theorem}[Proof in Appendix \ref{frankwolferegretappendix}]
\label{t:cgmregret} Let the step size be $\eta = \frac{1}{2n^{3/4}}$. Also let the mixing rates be $\gamma_t = \min\{1,2/t^{1/2}\}$, then Algorithm \ref{a:CG_algo} attains regret of $$\mathcal{R}_n  \le 8 \mathcal{G}^2n^{3/4}.$$
\end{theorem}

\section{Application: General Quadratic Losses} \label{sec:applications}
The first example of losses that we present are general quadratic losses. At each round the adversary can choose a symmetric (not necessarily positive semi-definite matrix) $A \in \mathbb{R}^{d \times d}$, and a vector $b\in \mathbb{R}^d$, with a constraint on the norm of the matrix and vector such that $\lVert A \rVert_{F}^2 + \lVert b \rVert_2^2 \le \mathcal{G}^2$. If we embed this pair into a Hilbert space defined by the feature map $(A,b)$ we get a kernel loss defined as -- $\langle \Phi(x) , (A,b) \rangle_{\mathcal{H}} = x^{\top}Ax+ b^{\top}x$, where $\Phi(x) = (xx^{\top},x)$ is the associated feature map for any $x \in \mathcal{A}$ and the inner product in the Hilbert space is defined as the concatenation of the trace inner product on the first coordinate and the Euclidean inner product on the second coordinate. The cumulative loss that the player aspires to minimize is, $\sum_{t=1}^n x_t^{\top} A_t x_t + b_t^{\top} x_t.$ The setting without the linear term, that is when $b_t = 0$ with positive semidefinite matrices $A_t$ is previously well studied in \citet{warmuth2006online,warmuth2008randomized,garber2015online,allen2017follow}. When the matrix is not positive semi-definite (making the losses non-convex) and there is a linear term, regret guarantees and tractable algorithms have not been studied even in the full information case. 

As this is a kernel loss we have regret bounds for these losses. We demonstrate in the subsequent sections in the full information case it is also possible to run our algorithms efficiently. First for exponential weights we show sampling is efficient for these losses.
\begin{lemma}[Proof in Appendix \ref{app:regretexp}]\label{lem:quadraefficient}
Let $B \in \mathbb{R}^{d \times d}$ be a symmetric matrix and $b \in \mathbb{R}^d$. Sampling from $q(a) \propto \exp( a^\top B a + a^\top b)$ for $\lVert a \rVert_2 \leq 1, a \in \mathbb{R}^d$ is tractable in $\tilde{\mathcal{O}}(d^4)$ time.
\end{lemma}
\subsection{Guarantees for Conditional Gradient Descent}
We now demonstrate that conditional gradient descent also can be run efficiently when the adversary plays a general quadratic loss. At each round the conditional gradient descent requires the player to solve the optimization problem, $v_t = \argmin_{a\in \mathcal{A}} \langle \nabla F_t(X_t), \Phi(a) \rangle_{\mathcal{H}}$. When the set of actions is $\mathcal{A} = {a \in \mathbb{R}^d : \lVert a \rVert_2 \le 1}$ then under quadratic losses this problem becomes,
\begin{align} \label{qcqp}
v_t &= \argmin_{a \in \mathcal{A}} a^{\top} B a + b^{\top}a,
\end{align}
for an appropriate matrix $B$ and $b$ that can be calculated by aggregating the adversary's actions up to step $t$. Observe that the optimization problem in Equation \eqref{qcqp} is a quadratically constrained quadratic program (QCQP) given our choice of $\mathcal{A}$. The dual problem is the (semi-definite program) SDP,
\begin{align*}
&max - t  - \mu \qquad \text{s. t.} \qquad \begin{bmatrix}
B + \mu I & b/2 \\
b/2 & t
\end{bmatrix} \succ 0.
\end{align*}
For this particular program with a norm ball constraint set it is known the duality gap is zero provided Slater's condition holds, that is, strong duality holds \citep[see Annex B.1][]{boyd2004convex}. 

Another example of losses where our framework is computationally efficient is when the underlying losses are posynomials (class of polynomials). We present this discussion in Appendix \ref{sec:posynomials}.
\section{Conclusions}
Under bandit feedback it would be interesting to explore if it is possible to establish lower bounds on the regret under the eigendecay conditions stated. Another interesting technical challenge is to see if Lemma \ref{lem:biascontrolbad} which we use to control the bias in our estimators can be sharpened to provide non-trivial regret guarantees even for slow eigendecay ($1<\beta\le 2$). Finding more kernel losses where our algorithms are provably computationally efficient is another direction that is exciting. Finally analyzing a mirror descent type algorithm under this framework could be useful to efficiently solve a wider class of problems. 
\subsubsection*{Acknowledgments}
We gratefully acknowledge the support of the NSF through grant IIS-1619362 and of the Australian Research Council through an Australian Laureate Fellowship (FL110100281) and through the Australian Research Council Centre of Excellence for Mathematical and Statistical Frontiers (ACEMS).

\nocite{*}
\bibliography{ref} 
\newpage
\appendix
\begin{flushleft}
\textbf{\Large Appendix}
\end{flushleft}
\subsection*{Organization of the Appendix and Roadmap of the Proof}
Here we describe the general organization of the proofs of the paper. We use the same notation for parameters as throughout the paper.

In Appendix \ref{app:regretboundbandit} we provide a proof of Theorem \ref{thm:mainregretbound} and Corollary \ref{cor:specificregretbound}. At a high level, the elements to prove this theorem are similar to that of proving regret bounds for linear losses. We first decompose the regret into an \emph{approximation error} term that arises due to the construction of the finite dimensional proxy $\Phi_m(\cdot)$ and another term which corresponds to the regret of a finite dimensional linear loss game (see Equation \ref{eq:epsilonbias}). To prove Theorem \ref{thm:mainregretbound} we then proceed in Appendix \ref{app:proofoftheorembandit} to control the regret of this finite dimensional linear bandit game by classical techniques. Crucially we also control terms that arise due to the bias in our estimators by invoking Lemma \ref{lem:biascontrolbad}.

In Appendix \ref{app:kernelPCA} we introduce and discuss ideas related to kernel principal component analysis (PCA). While in Appendix \ref{app:proxykernel} we prove Theorem \ref{thm:proxy_properties1}. Recall that this theorem was vital in establishing that the finite dimensional feature map we construct in Algorithm \ref{a:expweightskernels} induces a kernel $\kc$ that is an $\epsilon$-approximation of $\ke$. In Appendix \ref{app:covariancesample} we establish bounds on the sample complexity and control the error if the sample covariance matrix is used instead of the full covariance matrix in Algorithm \ref{a:expweightskernels}. 

The results about the full information setting, specifically the proofs of Theorems \ref{t:expweightsregret} and \ref{t:cgmregret} are provided in Appendix \ref{app:fullinfo}. In Appendix \ref{sec:posynomials} we apply our framework to posynomial losses, in Appendix \ref{sec:technicalstuff} we discuss Hoeffding's inequality and John's Theorem. Finally in Appendix \ref{sec:experiments} we present experimental evidence to verify our claims.
\section{Bandits Exponential Weights Regret Bound}
\label{app:regretboundbandit}
In this section we prove the regret bound stated in Section \ref{ss:banditregretalgo}. Here we borrow all the notation from Section \ref{sec:banditintro}. As defined before the expected regret for Algorithm \ref{a:expweightskernels} after $n$ rounds is
\begin{multline*}
\mathcal{R}_n  = \mathbb{E}\left[\sum_{t=1}^n \langle \Phi(a_t),w_t\rangle_{\mathcal{H}} - \langle \Phi(a^*),w_t\rangle_{\mathcal{H}}  \right]  = \mathbb{E}\left[ \sum_{t=1}^n \mathbb{E}_{a_t\sim p_t}\left[ \langle \Phi(a_t),w_t\rangle_{\mathcal{H}} - \langle \Phi(a^*),w_t\rangle_{\mathcal{H}} \Big \lvert \mathcal{F}_{t-1} \right] \right],
\end{multline*}
where $p_t$ is the exponential weights distribution described in Algorithm \ref{a:expweightskernels}, $a^*$ is the optimal action and $\mathcal{F}_{t-1}$ is the sigma field that conditions on ($a_1,a_2,\ldots,a_{t-1},y_1,y_2,\ldots,y_{t-1},y_t$), the events up to the end of round $t-1$. We will prove the regret bound for the case when the kernel is infinite dimensional, that is, the feature map $\Phi(a) \in \mathbb{R}^{D}$, where $D = \infty$. When $D$ is finite the proof is identical with $\epsilon = 0$. Recall that when $D$ is infinite we constrain the adversary to play rank-1 actions. We are going to refer to the adversarial action as $w_t =: \Phi(y_t)$ for some $y_t \in \mathbb{R}^d$. We now expand the definition of regret and get,
\begin{multline*}
\mathcal{R}_n  = \mathbb{E}\left[ \sum_{t=1}^n \mathbb{E}_{a_t\sim p_t}\left[ \langle \Phi(a_t),w_t\rangle_{\mathcal{H}} - \langle \Phi_m(a_t),w_t\rangle_{\mathcal{H}} \Big \lvert \mathcal{F}_{t-1} \right] \right] 
 \\ + \mathbb{E}\left[ \sum_{t=1}^n \mathbb{E}_{a_t\sim p_t}\left[ \langle \Phi_m(a^*),w_t\rangle_{\mathcal{H}} - \langle \Phi(a^*),w_t\rangle_{\mathcal{H}} \Big \lvert \mathcal{F}_{t-1} \right] \right]\\  + \underbrace{\mathbb{E}\left[ \sum_{t=1}^n \mathbb{E}_{a_t\sim p_t}\left[ \langle \Phi_m(a_t),w_t\rangle_{\mathcal{H}} - \langle \Phi_m(a^*),w_t\rangle_{\mathcal{H}} \Big \lvert \mathcal{F}_{t-1} \right] \right]}_{=:\mathcal{R}_n^m}.
\end{multline*}
Here $\mathcal{R}_n^m$ is the regret when we play the distribution in Algorithm \ref{a:expweightskernels} but are hit with losses that are governed by the kernel -- $\kc(\cdot,\cdot)$ (with the same $a^*$ as before). Observe that in $R_n^m$ only the component of $w_t$ in the subspace $\hat{S}_m$ contributes to the inner product, thus every term is of the form 
\begin{align*}
\langle \Phi_m(a_t),w_t \rangle & = \langle \Phi_m(a_t),\Phi_m(y_t) \rangle = \kc(y_t,a_t).
\end{align*}
As the proxy kernel $\kc$ is uniformly close by Theorem \ref{thm:proxy_properties1} we have,
\begin{align}\label{eq:epsilonbias}
\mathcal{R}_n & \le 2\varepsilon n + \mathcal{R}_n^m.
\end{align}

\subsection{Proof of Theorem \ref{thm:mainregretbound}} \label{app:proofoftheorembandit}
We will now attempt to bound $\mathcal{R}_n^m$ and prove Theorem \ref{thm:mainregretbound}. First we define the unbiased estimator (conditioned on $\mathcal{F}_{t-1}$) of $\Phi_m(y_t)$ at each round,
\begin{align}\label{eq:theunbiased}
\tilde{w}_t := \kc(a_t,y_t)  \left((\Sigma_m^{(t)})^{-1}\Phi_m(a_t)\right), \qquad t\in\{1,\ldots,n \},
\end{align}
where $\Phi_m(y_t) = P_{\hat{S}_m}(\Phi(y_t)) = P_{\hat{S}_m}(w_t)$. We cannot build $\tilde{w}_t$ as we do not receive $\kc(a_t,y_t)$ as feedback. Thus we also have
\begin{align*} 
&\mathbb{E}_{a_t \sim p_t}\left[ \hat{w}_t \lvert \mathcal{F}_{t-1}\right] = \mathbb{E}\left[\ke(a_t,y_t)\left((\Sigma_m^{(t)})^{-1}\Phi_m(a_t)\right)\Big\lvert \mathcal{F}_{t-1}\right] \\
 & =\Phi_m(y_t) + \mathbb{E}\left[\underbrace{\left(\ke(a_t,y_t)-\kc(a_t,y_t) \right) \left((\Sigma_m^{(t)})^{-1}\Phi_m(a_t)\right)}_{=:\xi_t, \text{ the bias}} \Big \lvert \mathcal{F}_{t-1}\right], t\in \{1,\ldots,n \}. \numberthis \label{eq:biasadversary}
\end{align*}
If $\Phi_m(\cdot) = \Phi(\cdot)$ then the bias $\xi_t$ would be zero. 
We now present some estimates involving $\tilde{w}_t$. In the following section we sometimes denote $\tilde{w}_t$ and $\hat{w}_t$ more explicitly as $\tilde{w}_t(a_t)$ and $\hat{w}_t(a_t)$ where there may be room for confusion.
\begin{lemma} \label{biaslem} For any fixed $a \in \mathcal{A}$ we have,
\begin{align*}
\mathbb{E}_{a_t \sim p_t}\left[\langle \tilde{w_t}(a_t),\Phi_m(a)\rangle\Big\lvert \mathcal{F}_{t-1} \right] & = \kc(y_t,a) , \qquad t \in \{1,\ldots,n \}. 
\end{align*}
We also have for all $t \in \{1,\ldots,t \}$,
\begin{align*}
\mathbb{E}_{a_t\sim p_t}\left[ \kc(y_t,a_t) \Big \lvert \mathcal{F}_{t-1}\right]   = \mathbb{E}_{a_t\sim p_t}\left[\sum_{a \in \mathcal{A}}p_t(a)\left\langle \tilde{w}_t(a_t) , \Phi_m(a) \right\rangle  \Big\rvert \mathcal{F}_{t-1}  \right] .
\end{align*}
\end{lemma}
\begin{Proof} The first claim follows by Equation \eqref{eq:theunbiased} and the linearity of expectation we have
\begin{align*}
\mathbb{E}_{a_t \sim p_t} \left[\langle \tilde{w}_t(a_t),\Phi_m(a)\rangle\Big \lvert \mathcal{F}_{t-1} \right]  = \langle \mathbb{E}\left[ \tilde{w}_t(a_t) \lvert \mathcal{F}_{t-1}\right],\Phi_m(a) \rangle 
 = \kc(y_t,a).
\end{align*}
where the expectation is taken over $p_t$. Now to prove the second part of the theorem statement we will use tower property. Observe that conditioned on $\mathcal{F}_{t-1}$, $p_t$ and $a_t$ are measurable.
\begin{align*}
\mathbb{E}\left[\kc(y_t,a_t) \Bigg \lvert \mathcal{F}_{t-1} \right] 
 &= \mathbb{E}\left[\mathbb{E}_{a \sim p_t}\left[ \langle \tilde{w}_t(a_t),\Phi_m(a) \rangle  \Bigg \lvert a_t\right]\Bigg\lvert \mathcal{F}_{t-1}\right] = \mathbb{E}\left[\sum_{a \in \mathcal{A}}p_t(a)\left\langle \tilde{w}_t (a_t), \Phi_m(a) \right\rangle   \Big\rvert \mathcal{F}_{t-1} \right].
\end{align*}
\end{Proof}
We are now ready to prove Theorem \ref{thm:mainregretbound} and establish the claimed regret bound.
\begin{Proof}[Proof of Theorem \ref{thm:mainregretbound}] The proof is similar to the regret bound analysis of exponential weights for linear bandits. We proceed in 4 steps. In the first step we decompose the cumulative loss in terms of an exploration cost and an exploitation cost. In Step 2 we control the exploitation cost by using Hoeffding's inequality as is standard in linear bandits literature, but additionally we need to control terms arising out of the bias of our estimate. In Step 3 we bound the exploration cost and finally in the fourth step we combine the different pieces and establish the claimed regret bound.

\textbf{Step 1:} Using Lemma \ref{biaslem} and the fact that $\tilde{w}_t$ is an unbiased estimate of $\Phi_m(y_t)$ we can decompose the cumulative loss, the first term in $R_n^m$ as
\begin{multline*}
\mathbb{E}\left[\sum_{t=1}^{n} \kc(a_t,y_t) \right]  = \mathbb{E}\left[ \sum_{t=1}^n \left[ \sum_{a \in \mathcal{A}} p_t(a) \left \langle \tilde{w}_t(a_t), \Phi_m(a) \right \rangle \Big\rvert  \mathcal{F}_{t-1}\right]\right]  \\
\numberthis \label{eq:masterregret} = \underbrace{(1-\gamma)\mathbb{E}\left[ \sum_{t=1}^n \left[ \sum_{a \in \mathcal{A}}q_t(a) \left \langle \tilde{w}_t, \Phi_m(a) \right \rangle \Big\rvert  \mathcal{F}_{t-1}\right]\right]}_{Exploitation}   + \underbrace{\gamma \cdot \mathbb{E}\left[ \sum_{t=1}^n \left[ \sum_{a \in \mathcal{A}}\nu_{\mathcal{J}}^{\mathcal{A}}(a) \left \langle \tilde{w}_t, \Phi_m(a) \right \rangle \Big\rvert  \mathcal{F}_{t-1}\right]\right]}_{Exploration},
\end{multline*}
where the second line follows by the definition of $p_t$. 

\textbf{Step 2:} We first focus on the `Exploitation' term. 
\begin{align*}
&\clubsuit := (1-\gamma)\mathbb{E}\left[ \sum_{t=1}^n \left[ \sum_{a \in \mathcal{A}}q_t(a) \left \langle \tilde{w}_t, \Phi_m(a) \right \rangle \Big\rvert  \mathcal{F}_{t-1}\right]\right] \\& = (1-\gamma)\underbrace{\mathbb{E}\left[ \sum_{t=1}^n \left[ \sum_{a \in \mathcal{A}}q_t(a) \left \langle \hat{w}_t, \Phi_m(a) \right \rangle \Big\rvert  \mathcal{F}_{t-1}\right]\right]}_{=:\spadesuit} + (1-\gamma)\underbrace{\mathbb{E}\left[ \sum_{t=1}^n \left[ \sum_{a \in \mathcal{A}}q_t(a) \left \langle \tilde{w}_t-\hat{w}_t, \Phi_m(a) \right \rangle \Big\rvert  \mathcal{F}_{t-1}\right]\right]}_{=:\diamondsuit}. \numberthis \label{eq:exploitationdecompositon} 
\end{align*}
 Under our choice of $\gamma$ by Lemma \ref{l:boundonlossestimate} (proved in Appendix \ref{ss:techresultsregret}) we know that $\eta \langle \hat{w}_t, \Phi(a_t) \rangle > -1$. Therefore by Hoeffding's inequality (Lemma \ref{hff}) we get,
\begin{align}
\nonumber\spadesuit = \mathbb{E}\left[\sum_{t=1}^{n} \left[\sum_{a \in \mathcal{A}}  q_t(a) \left \langle \hat{w}_t, \Phi_m(a) \right \rangle  \Big \lvert \mathcal{F}_{t-1} \right] \right]& \le -\frac{1}{\eta} \underbrace{\mathbb{E}\left[\sum_{t=1}^{n} \log\left( \mathbb{E}_{a\sim q_t}\left[ \exp\left(-\eta \left \langle \hat{w}_t(a_t), \Phi_m(a) \right \rangle \right)\right]\right)\right]}_{=:\Gamma_1}\\  &  + \underbrace{(e-2) \eta \mathbb{E}\left[\sum_{t=1}^n \left[\sum_{a \in \mathcal{A}} q_t(a)\left( \left \langle \hat{w}_t, \Phi_m(a) \right \rangle\right)^2 \Big \lvert \mathcal{F}_{t-1} \right] \right]}_{=:\Gamma_2}.\label{eq:gammadefinitions}
\end{align}
Both $\Gamma_1$ and $\Gamma_2$ can be bounded by standard techniques established in the literature of adversarial linear bandits. We will see that $\Gamma_1$ is a telescoping series and is controlled in Lemma \ref{l:gamma1control}. While the second term $\Gamma_2$ is the variance of the estimated loss is bounded in Lemma \ref{l:variancecontrol}. We defer the proof of both Lemma \ref{l:gamma1control} and Lemma \ref{l:variancecontrol} to Appendix \ref{ss:techresultsregret}. Plugging in the bounds on $\Gamma_1$ and $\Gamma_2$ into Equation \eqref{eq:gammadefinitions} we get,
\begin{align*}\numberthis\label{eq:exploitcontrol}
&(1-\gamma)\spadesuit = (1-\gamma)\cdot\mathbb{E}\left[\sum_{t=1}^{n} \left[\sum_{a \in \mathcal{A}}  q_t(a) \left \langle \hat{w}_t, \Phi_m(a) \right \rangle  \Big \lvert \mathcal{F}_{t-1} \right] \right] \\&\le  \mathbb{E}\left[\sum_{i=1}^{n}\kc(a^*,y_t) \right] + \frac{1}{\eta}\log\left(  \lvert \mathcal{A} \rvert \right) + \mathbb{E}\left[\sum_{t=1}^n \langle \hat{w}_t-\tilde{w}_t,\Phi_m(a^*) \rangle \right] + (e-2)\eta \mathcal{G}^4 m n \\
& \le \mathbb{E}\left[\sum_{i=1}^{n}\kc(a^*,y_t) \right] + \frac{1}{\eta}\log\left(  \lvert \mathcal{A} \rvert \right) + \frac{\epsilon n}{\mathcal{G}^2\eta} + (e-2)\eta \mathcal{G}^4 m n,
\end{align*}
where the last inequality follows by Lemma \ref{lem:biascontrolbad}. Also by Lemma \ref{lem:biascontrolbad} we get, $\diamondsuit \le \epsilon n/\mathcal{G}^2\eta$. Combining this with Equation \eqref{eq:exploitationdecompositon} we get,
\begin{align} \label{eq:exploitmasterbound}
\clubsuit & \le \mathbb{E}\left[\sum_{i=1}^{n}\kc(a^*,y_t) \right] + \frac{1}{\eta}\log\left(  \lvert \mathcal{A} \rvert \right) + \frac{2\epsilon n}{\mathcal{G}^2\eta} + (e-2)\eta \mathcal{G}^4 m n.
\end{align}

\textbf{Step 3:} Next, observe that the exploration term is bounded above as
\begin{align}\label{}
\gamma \cdot \mathbb{E}\left[ \sum_{t=1}^n \mathbb{E}\left[ \sum_{a \in \mathcal{A}}\nu_{\mathcal{J}}^{\mathcal{A}}(a) \left \langle \tilde{w}_t, \Phi_m(a) \right \rangle \Big\rvert  \mathcal{F}_{t-1}\right]\right] & \le 4\gamma \mathcal{G}^2 n,
\end{align}
where the above inequality follows by Lemma \ref{biaslem} and Cauchy-Schwartz inequality along with the fact that $\epsilon \le \mathcal{G}^2$.

\textbf{Step 4:} Putting these all these together into Equation \eqref{eq:masterregret} we get the desired bound on the finite dimensional regret
\begin{align*}
\mathcal{R}_n^m = \mathbb{E} \left[\sum_{t=1}^n \left( \kc(a_t,y_t) - \kc(a^*,y_t) \right) \right] & \le 4 \gamma \mathcal{G}^2n + (e-2) \mathcal{G}^4\eta m n + \frac{2\epsilon n}{\mathcal{G}^2\eta} + \frac{1}{\eta} \log(\lvert \mathcal{A} \rvert).
\end{align*}
Plugging the above bound on $\mathcal{R}_n^m$ into Equation \eqref{eq:epsilonbias} we get a bound on the expected regret as
\begin{align*}
\mathcal{R}_n \le 4\gamma \mathcal{G}^2 n + (e-2) \mathcal{G}^4\eta m n + 2\epsilon n + \frac{2\epsilon n}{\mathcal{G}^2\eta}+\frac{1}{\eta} \log(\lvert \mathcal{A} \rvert),
\end{align*}
completing the proof.
\end{Proof}
\subsubsection{Technical Results used in Proof of Theorem \ref{thm:mainregretbound}} \label{ss:techresultsregret}
First let us focus on bounding $\Gamma_1$. A term analogous to $\Gamma_1$ also appears in the regret bound analysis for exponential weights in the adversarial linear bandits setting; we adapt those proof techniques here to work with biased estimates ($\hat{w}_t$). 
\begin{lemma}\label{l:gamma1control}
Let $\Gamma_1$ be as defined in Equation \eqref{eq:gammadefinitions} then we have
\begin{align*}
\Gamma_1 & \ge -\eta\mathbb{E}\left[ \sum_{i=1}^{n}\kc(a^*,y_t)\right] -\log\left(  \lvert \mathcal{A} \rvert \right) -\eta\mathbb{E}\left[\sum_{t=1}^n \langle \hat{w}_t-\tilde{w}_t,\Phi_m(a^*) \rangle \right] .
\end{align*}
\end{lemma}
\begin{Proof} Expanding $\Gamma_1$ we get
\begin{align*}
\Gamma_1 & = \mathbb{E}\left[\sum_{t=1}^{n} \log \left(\mathbb{E}_{a\sim q_t}\left[ \exp\left(-\eta \left \langle \hat{w}_t, \Phi_m(a) \right \rangle \right)\right] \right) \right]\\ 
& \overset{(i)}{=} \mathbb{E}\left[\sum_{t=1}^{n} \log\left\{ \frac{\sum_{a \in \mathcal{A}}\exp\left( -\eta \sum_{i=1}^{t-1}\left \langle \hat{w}_i, \Phi_m(a) \right \rangle\right)\cdot\exp\left( -\eta \left \langle \hat{w}_t, \Phi_m(a) \right \rangle\right)}{ \sum_{a \in \mathcal{A}} \exp\left( -\eta \sum_{i=1}^{t-1}\left \langle \hat{w}_i, \Phi_m(a) \right \rangle\right)} \right\}\right]\\
& \overset{(ii)}{=} \mathbb{E}\left[\log\left(\sum_{a \in \mathcal{A}}\exp\left( -\eta \sum_{i=1}^{n}\left \langle \hat{w}_i, \Phi_m(a) \right \rangle\right) \right) -\log\left( \lvert \mathcal{A}\rvert\right) \right],\numberthis \label{eq:gamma1masterbound}
\end{align*}
where $(i)$ follows by the definition of $q_t(a)$ and $(ii)$ follows as the sum telescopes and we start of with the uniform distribution over $\mathcal{A}$. We have for any element $a'\in \mathcal{A}$,
\begin{align*}
\mathbb{E}\left[\log\left(\sum_{a \in \mathcal{A}}\exp\left( -\eta \sum_{i=1}^{n}\left \langle \hat{w}_i, \Phi_m(a) \right \rangle\right) \right) \right]&\ge -\mathbb{E}\left[\eta \sum_{i=1}^{n}\left \langle \hat{w}_i, \Phi_m(a') \right \rangle\right]\\ 
& = -\eta \mathbb{E}\left[\sum_{t=1}^n \langle \tilde{w}_t,\Phi_m(a') \rangle \right] - \eta\mathbb{E}\left[\sum_{t=1}^n \langle \hat{w}_t-\tilde{w}_t,\Phi_m(a') \rangle \right] \\
& = -\eta \mathbb{E}\left[\sum_{t=1}^n \kc(a',y_t) \right] -\eta\mathbb{E}\left[\sum_{t=1}^n \langle \hat{w}_t-\tilde{w}_t,\Phi_m(a') \rangle \right],
\end{align*}
where the last equality by Lemma \ref{biaslem}. Choosing $a' = a^*$ and plugging this lower bound into Equation \eqref{eq:gamma1masterbound} completes the proof.
\end{Proof}

The next lemma controls of the variance of the expected loss -- $\Gamma_2$. A term analogous to $\Gamma_2$ appears in the regret bound analysis of exponential weights in adversarial linear bandits which we adapt to our setting.
\begin{lemma} \label{l:variancecontrol}
Let $\Gamma_2$ be defined as in Equation \eqref{eq:gammadefinitions} and the choice of parameters as specified in Theorem \ref{thm:mainregretbound} then we have
\begin{align}
\Gamma_2 = (e-2) \eta \mathbb{E}\left[\sum_{t=1}^n \mathbb{E}_{a \sim q_t}\left[\left \langle \hat{w}_t,\Phi_m(a) \right\rangle^2 \Big\lvert \mathcal{F}_{t-1}\right] \right] & \le (e-2)\mathcal{G}^4\eta m n/(1-\gamma).
\end{align}
\end{lemma}
\begin{Proof} Note that by definition of $q_t$, we have that $(1-\gamma)q_t(a) \le p_t(a)$. To ease notation let $\Sigma_t := \Sigma_m^{(t)} = \mathbb{E}_{p_t}\left[\Phi_m(x) \Phi_m(x)^{\top} \right]$. Taking expectation over the randomness in $\hat{w}_t$ for any fixed $a$ we have,
\begin{align*}
\mathbb{E}_{a_t\sim q_t}\left[ \langle \hat{w}_t(a_t), \Phi_m(a) \rangle^2 \right] & = \Phi_m(a)^{\top} \mathbb{E}\left[\hat{w}_t \hat{w}_t^{\top} \right] \Phi_m(a) \\
& = \Phi_m(a)^{\top} \mathbb{E}_{a_t \sim p_t}\left[ \ke(a_t,y_t)^2 \Sigma_t^{-1} \Phi_m(a_t)\Phi_m(a_t)^{\top} \Sigma_t^{-1}\right] \Phi_m(a)\\
& \le \mathcal{G}^4\Phi_m(a)^{\top} \Sigma_t^{-1} \Phi_m (a).
\end{align*}
where the second equality follows by the definition of $\hat{w}_t$. Given this calculation we now also take expectation over the choice of $a$ so we have,
\begin{align*}
\mathbb{E}_{a\sim p_t} \left[\mathbb{E}_{a_t \sim p_t} \left[ \langle \hat{w}_t, \Phi_m(a) \rangle^2\right] \right] & \le \mathcal{G}^4\mathbb{E}_{a \sim p_t} \left[ tr\left(\Phi_m(a)^{\top} \Sigma_t^{-1} \Phi_m(a)\right)\right] \\
& = \mathcal{G}^4tr \left( \Sigma_t^{-1} \mathbb{E}_{a \sim p_t} \left[ \Phi_m(a) \Phi_m(a)^{\top} \right]\right) = \mathcal{G}^4tr\left(I_{m \times m}\right) = \mathcal{G}^4m.
\end{align*}
Summing over $t=1$ to $n$ establishes the result.
\end{Proof}
We now prove the bound on the terms that arise out of our biased estimates.
\begin{lemma}\label{lem:biascontrolbad} Let $\gamma =4\eta \mathcal{G}^4m$  and let $\epsilon \le \mathcal{G}^2$, then for all $a\in \mathcal{A}$ and for all $t\in \{ 1,\ldots, n\}$ we have
\begin{align*}
\lvert \langle \hat{w}_t - \tilde{w}_t,\Phi_m(a) \rangle \rvert & \le \frac{\epsilon}{\eta \mathcal{G}^2}.
\end{align*}
\end{lemma}
\begin{Proof} By the definition of $\tilde{w}_t$ and $\hat{w}_t$,
\begin{align*}
\lv\tilde{w}_t - \hat{w}_t \rv_2 &= \left\lv \left(\kc(y_t,a_t) - \ke(y_t,a_t)\right) \left(\Sigma_m^{(t)}\right)^{-1}\Phi_m(a_t) \right\rv_2 \\
& = \lvert \kc(y_t,a_t) - \ke(y_t,a_t)\rvert\left\lv  \left(\Sigma_m^{(t)}\right)^{-1}\Phi_m(a_t) \right\rv_2 \\
& \le \epsilon \cdot \frac{m}{\gamma} \cdot (\mathcal{G}+\sqrt{\epsilon}),
\end{align*}
where the inequality follows as $\kc$ is an $\epsilon$-approximation of $\ke$, the minimum eigenvalue of $\Sigma_m^{(t)}$ is $\gamma/m$ by Proposition \ref{l:johneigenvaluebound}. So by Cauchy-Schwartz we get,
\begin{align*}
\lvert \langle \hat{w}_t - \tilde{w}_t,\Phi_m(a) \rangle \rvert \le \lv\tilde{w}_t - \hat{w}_t \rv_2\lv \Phi_m\rv_2 &\le \frac{\epsilon m (\mathcal{G}+\sqrt{\epsilon})^2}{\gamma},
\end{align*}
the claim now follows by the choice of $\gamma$ and by the condition on $\epsilon$.
\end{Proof}
While using Hoeffding's inequality to arrive at Inequality \eqref{eq:gammadefinitions} we assume that the estimate of the loss is lower bounded by $-1/\eta$. The next lemma help us establish that under the choice of $\gamma$ the exploration parameter in Theorem \ref{thm:mainregretbound} this condition holds.

\begin{lemma}\label{l:boundonlossestimate} Let $\epsilon \le \mathcal{G}^2$ then for any $a \in \mathcal{A}$ and for all $t=1,\ldots,n$ we have
\begin{align*} 
\lvert \langle \hat{w}_t,\Phi_m(a) \rangle \rvert \le \mathcal{G}^2 \sup_{c,d \in \mathcal{A}} \left \lvert \Phi_m(c)^{\top} \left( \mathbb{E}_{a \sim p_t} \left[ \Phi_m(a) \Phi_m(a)^{\top} \right]\right)^{-1} \Phi_m(d)\right \rvert.
\end{align*}
Further if the exploration parameter is $\gamma > 4\eta\mathcal{G}^4m$ then we have a bound on the estimated loss at each round
\begin{align*}
\eta\lvert \langle \hat{w}_t,\Phi_m(a) \rangle \rvert \le 1, & \qquad \forall a \in \mathcal{A}.
\end{align*}
\end{lemma}
\begin{Proof} Recall the definition of $\Sigma_m^{(t)} = \mathbb{E}_{p_t} \left[ \Phi_m(a) \Phi_m(a)^{\top}\right]$ (we drop the index $t$ to lighten notation in this proof). The proof follows by plugging in the definition of the loss estimate $\hat{w}_t$,
\begin{align*}
\lvert \hat{w}_t^{\top} \Phi_m(a) \rvert & =  \left \lvert \ke(a_t,y_t)  \left( \Sigma^{-1}_m \Phi_m(a_t)\right)^{\top}\Phi_m(a) \right \rvert \\ 
& \le \underbrace{\lvert \ke(a_t,y_t) \rvert}_{\le \mathcal{G}^2} \left \lvert  \left( \Sigma^{-1}_m \Phi_m(a_t)\right)^{\top}\Phi_m(a)\right \rvert \\ 
& \le 4\mathcal{G}^2\sup_{c,d \in \mathcal{A}} \left \lvert \Phi_m(c)^{\top} \Sigma_{m}^{-1} \Phi_m(d)\right \rvert. \numberthis \label{eq:supnormbound}
\end{align*}
Now note that the matrix $\Sigma_m$ has its lowest eigenvalue lower bounded by $\gamma/m$ by Proposition \ref{l:johneigenvaluebound} \citep[see also discussion by][]{bubeck2012towards}. Thus we have,
\begin{align*}
\sup_{c,d \in \mathcal{A}} \left \lvert \Phi_m(c)^{\top} \Sigma_{m}^{-1} \Phi_m(d)\right \rvert  & \le \frac{4\mathcal{G}^2 m}{\gamma},
\end{align*}
where the inequality above follows by the assumption that $\epsilon \le \mathcal{G}^2$. Combing this with Equation \eqref{eq:supnormbound} yields the desired claim.
\end{Proof}
\subsection{Proof of Corollary \ref{cor:specificregretbound}}
In this section we present the proof of Corollary \ref{cor:specificregretbound}, which establishes the regret bound under particular conditions on the eigen-decay of the kernel.
\begin{Proof}[Proof of Corollary \ref{cor:specificregretbound}] Given our assumption $\mathcal{G}=1$ and by the choice of $\gamma = 4\eta \mathcal{G}^4m$ the regret bound becomes,
\begin{align*}
\mathcal{R}_n \le \underbrace{20\eta m n }_{=:R_1}  +\underbrace{\frac{2\epsilon n}{\eta}}_{=: R_2} +\underbrace{\frac{1}{\eta} \log\left( \lvert \mathcal{A} \rvert \right)}_{=:R_3}+ \underbrace{2\epsilon n}_{=: R_4}.
\end{align*}
\textbf{Case 1:} First we assume $(C,\beta)$-polynomial eigen-value decay. By the results of Proposition \ref{prop:mbounds} we have a sufficient condition on the choice of $m$ for $\kc$ to be an $\epsilon$-approximation of $\ke$,
\begin{align*}
m = \left [\frac{4C\mathcal{B}^2}{(\beta-1)\epsilon}\right]^{1/\beta-1}.
\end{align*}
With this choice of $m$ we equate the terms $R_1,R_2$ and $R_3$ with each other. This yields the choice, 
\begin{align*}
\epsilon = \frac{\log(\lvert \mathcal{A} \rvert)}{2n}, \qquad \text{and}\qquad \eta^2 = \frac{\epsilon}{10 m}.
\end{align*}
Note that under this choice there exists a constant $n_0(\beta,C,\mathcal{B},\log(\lvert \mathcal{A} \rvert))>\log(\lvert \mathcal{A} \rvert)/2$  such that when $n>n_0$ then, $R_4< R_1$. Also note when $n > \log(\lvert A \rvert)/2$ then $\epsilon < 1 = \mathcal{G}^2$ so the conditions of Theorem \ref{thm:mainregretbound} are indeed satisfied. Plugging in these choice of $\epsilon,m$ and $\eta$ for $n > n_0$ yields,
\begin{align*}
\mathcal{R}_n \le 4 R_1 = \sqrt{160}\cdot \left[\frac{2^{\beta+2}C\mathcal{B}^2}{\beta-1}\right]^{\frac{1}{2(\beta-1)}}\cdot \left(\log(\lvert A \rvert)\right)^{\frac{\beta-2}{2(\beta-1)}}\cdot n^{\frac{\beta}{2(\beta-1)}}.
\end{align*}

\textbf{Case 2:} Here we assume $(C,\beta)$-exponential eigen-value decay. Again by the results of Proposition \ref{prop:mbounds} we have a sufficient condition for the choice of $m$ for $\kc$ to be an $\epsilon$- approximation of $\ke$,
\begin{align*}
m = \frac{1}{\beta}\log\left(\frac{4C \mathcal{B}^2}{\beta\epsilon}\right).
\end{align*}
Again as before, by equating $R_1,R_2,R_3$ yields $\epsilon = \log(\lvert \mathcal{A} \rvert)/(2n)$ and $\eta^2 = \epsilon/10 m$.
Again as with Case 1, there exists a constant $n_0(\beta,C,\mathcal{B},\log(\lvert \mathcal{A} \rvert))>\log(\lvert \mathcal{A} \rvert)/2$  such that when $n>n_0$ then, $R_4< R_1$. Plugging in these choice of $\epsilon,m$ and $\eta$ for $n > n_0$ yields,
\begin{align*}
\mathcal{R}_n \le 4R_1 = \sqrt{\frac{320\cdot\log(\lvert \mathcal{A} \rvert)}{\beta}\cdot \log\left( \frac{40C\mathcal{B}^2 n }{\beta \log(\lvert \mathcal{A} \rvert)}\right)\cdot  n}.
\end{align*}
\end{Proof}

\section{Kernel principal component analysis}
\label{app:kernelPCA}
We review the basic principles underlying kernel principal component analysis (PCA). Let $\ke$ be some kernel defined over $\mathcal{A} \subset \mathbb{R}^d$ and $x_1, \cdots, x_p \sim \mathbb{P}$ a probability measure over $\mathcal{A}$. Let us denote a feature map of $\ke$ by $\Phi : \mathbb{R}^d \rightarrow \mathbb{R}^D$.

The goal of PCA is to extract a set of eigenvalues and eigenvectors from a sample covariance matrix. In kernel PCA we want to calculate the eigenvectors and eigenvalues of the sample \emph{kernel} covariance matrix,
\begin{align*}
\hat{\Sigma} = \frac{1}{p} \sum_{i=1}^{p} \Phi(x_i)\Phi(x_i)^{\top}.
\end{align*}
When working in a reproducing kernel Hilbert space $\mathcal{H}$ in which no feature map is explicitly available, an alternative approach is taken by working instead with the sample Gram matrix.
\begin{lemma} \label{lem:sameeigenvalues}
Let $\Phi(x_1), \cdots, \Phi(x_p)$ be $p$ points in $\mathcal{H}$. The eigenvalues of the sample covariance matrix, $\frac{1}{p} \sum_{i=1}^p \Phi(x_i)\Phi(x_i)^\top$ equal the eigenvalues of the sample Gram matrix $\mathbb{K} \in \mathbb{R}^{p\times p}$, where the sample Gram matrix is defined entry-wise as $\mathbb{K}_{ij} = \frac{\ke(x_i, x_j)}{p}$.
\end{lemma}
\begin{Proof}
Let $X \in \mathbb{R}^{p\times D}$ be such that the $i^{th}$ row is $\frac{\Phi(x_i)}{\sqrt{p}}$. The singular value decomposition (SVD) of $X$ is 
\begin{align*}
X = U D V^\top,
\end{align*}
with $U \in \mathbb{R}^{p \times p}$, $D \in \mathbb{R}^{p \times D}$ and $V \in \mathbb{R}^{D \times D}$. Therefore $X^\top X = V D^\top D V^\top$ and $XX^\top = U DD^\top U^\top$. We identify $X^{\top}X$ as the sample covariance matrix and $XX^{\top}$ as the sample Gram matrix. Since $DD^\top$ and $D^\top D$ are both diagonal and have the same nonzero values this establishes the claim.
\end{Proof}

Another insight used in kernel PCA procedures is the observation that the span of the eigenvectors corresponding to nonzero eigenvalues of the sample covariance matrix $\frac{1}{p} \sum_{i=1}^p \Phi(x_i)\Phi(x_i)^\top$ is a subspace of the span of the data-points $\{ \Phi(x_i) \}_{i=1}^p$. This means that any eigenvector $v$ corresponding to a nonzero eigenvalue for the second moment sample covariance matrix can be written as a linear combination of the $p-$datapoints, $v_i = \sum_{j=1}^p \omega_{ij} \Phi(x_j)$ ($\omega_{ij}$ denotes the $j^{th}$ component of $\omega_i \in \mathbb{R}^p$). Observe that $v_i$ are the eigenvectors of the sample covariance matrix, so we have
\begin{equation*}
\left[ \frac{1}{p} \sum_{i=1}^p \Phi(x_i)\Phi(x_i)^\top \right] \left( \sum_{j=1}^p \omega_{ij} \Phi(x_j)   \right) = \mu_i \sum_{j=1}^p \omega_{ij} \Phi(x_j).
\end{equation*}
This implies we may consider solving the equivalent system
\begin{equation}\label{eq:kernel_PCA_eq1}
\mu_i \left( \langle \Phi(x_k) , v_i \rangle_{\mathcal{H}} \right) = \left \langle \Phi(x_k), \left(\frac{1}{p} \sum_{i=1}^p \Phi(x_i)\Phi(x_i)^\top  \right)  v_i  \right \rangle_{\mathcal{H}} \text{  } \forall k = 1, \cdots, p.
\end{equation}

Substituting $v_ i = \sum_{j=1}^p \omega_{ij} \Phi(x_j)$ into Equation \eqref{eq:kernel_PCA_eq1}, and using the definition of $\mathbb{K}$ we obtain
\begin{equation*}
\mu_i \mathbb{K} \omega_i = \mathbb{K}^2 \omega_i.
\end{equation*}
To find the solution of this last equation we solve the eigenvalue problem,
\begin{equation*}
\mathbb{K}\omega_i = \mu_i \omega_i.
\end{equation*}
Once we solve for $\alpha_i$ we can recover the eigenvector of the sample covariance matrix by setting $v_ i = \sum_{j=1}^p \omega_{ij} \Phi(x_j)$.

\section{Proxy Kernel Properties}\label{app:proxykernel}
In this section we prove Theorem \ref{thm:proxy_properties1}. We reuse the notation introduced in Section \ref{sec:constrproxy} which we recall here. 

Let $\{ \mu_j \}_{j=1}^\infty$ be the Mercer's eigenvalues of a kernel $\ke$ under measure $\mathbb{P}$ with eigenfunctions $\{ \phi_j \}_{j=1}^\infty$, and we assume that $\sup_{j\in \mathbb{N}} \sup_{x \in \mathcal{A}} \lvert \phi_j(x) \rvert \leq \mathcal{B}$ for some $\mathcal{B} < \infty$. Let $m(\epsilon)$ be such that $\sum_{j=m+1}^\infty \mu_j \leq \frac{\epsilon}{4\mathcal{B}^2}$ and denote the $m^{th}$ eigen-gap as $\delta_m = \frac{1}{2}\left( \mu_m - \mu_{m+1}  \right)$. Denote by $S_m$ and $\hat{S}_m$ the subspaces spanned by the first $m$ eigenvectors of the covariance matrix $\mathbb{E}_{x \sim \mathbb{P}}\left[ \Phi(x) \Phi(x)^\top  \right]$ and a sample covariance matrix $\frac{1}{p}\sum_{i=1}^p \Phi(x_i) \Phi(x_i)^\top$ respectively. Define $P_{S_m}$ and $P_{\hat{S}_m}$ to be the projection operators to $S_m$ and $\hat{S}_m$. Recall the definition of 
\begin{align*}
\hat{\ke}^o_m(x,y) = \langle P_{S_m} (\Phi(x)), P_{S_m} (\Phi(y)) \rangle_\mathcal{H} = \langle \Phi_m^o(x),\Phi_m^o(y) \rangle_{\mathcal{H}},
\end{align*}
a deterministic approximate kernel and the stochastic proxy approximate kernel 
\begin{align*}
\kc( x,y) = \langle P_{\hat{S}_m} (\Phi(x)), P_{\hat{S}_m} (\Phi(y)) \rangle_\mathcal{H},
\end{align*} 
with associated feature map $\Phi_m(x) = P_{\hat{S}_m} \Phi(x)$. We first prove Lemma \ref{lemma_chopped_kernel} restated here.
\detelemma*
\begin{Proof}
By definition, for all $x,y \in \mathcal{A}$
\begin{align*}
     \ke(x,y) - \kc^o( x,y)  &= \sum_{j=m+1}^\infty \mu_j \phi_j(x) \phi_j(y)  \leq  \sum_{j=m+1}^\infty \mu_j \lvert\phi_j(x) \phi_j(y) \rvert   \leq  \sum_{j=m+1}^\infty \mu_j \mathcal{B}^2 \leq \frac{\epsilon}{4}.
\end{align*}
The reverse inequality; $\kc^o(x,y) - \ke(x,y) \leq \frac{\epsilon}{4}$, is also true therefore,
\begin{equation*}
   \lvert \ke(x,y) - \kc^o( x,y) \rvert \leq \frac{\epsilon}{4},
\end{equation*}
for all $x,y \in \mathcal{A}$.
\end{Proof}

We now state and prove an expanded version of Theorem \ref{thm:proxy_properties1} (where $w = \min(\sqrt{\epsilon}/2,\delta_m/2)$) which is used to establish the $\epsilon$-approximability of the stochastic kernel $\kc$.
\begin{theorem}\label{thm:proxy_properties}
Let $\epsilon,m, \mathbb{P}$ be as in Lemma \ref{lemma_chopped_kernel}. Define the $m$-th level eigen-gap as $\delta_m = \frac{1}{2}\left( \mu_m - \mu_{m+1}  \right)$. Also let $B_m = \frac{2\mathcal{G}^2}{\delta_m} \left( 1 + \sqrt{ \frac{\alpha}{2}}   \right)$, $\delta_m/2>w>0$  and $p \geq \frac{B_m^2 \mathcal{G}^2 }{w^2} $. The finite dimensional proxies $\hat{\ke}_m^o$ and $\hat{\ke}_m$ satisfy the following properties with probability $1-e^{-\alpha}$:
\begin{enumerate}
\item $|\ke(x,y) - \hat{\ke}_m(x,y) | \leq \frac{\epsilon}{4} + \sqrt{\epsilon}w + w^2$.
\item $|\hat{\ke}_m(x,y) - \hat{\ke}_m^o(x,y) | \leq w^2, \qquad \qquad \qquad \forall$ $x,y \in \mathcal{A}$.
\item The Mercer operator eigenvalues $\mu^{(m)}_1 \geq \cdots \geq \mu^{(m)}_m$ and $\mu_1 \geq \cdots \geq \mu_m$ of $\kc$ and $\kc^o$ follow $
\sup_{i=1, \cdots, m} |\mu^{(m)}_i - \mu_i| \leq w^2$.
\end{enumerate}
\end{theorem}
Theorem \ref{thm:proxy_properties} shows that, as long as sufficiently samples $p(m)$ are used, with high probability $\kc$ is uniformly close to $\hat{\ke}_m^o$ and therefore to $\ke$. We prove this theorem by a series of lemmas and auxiliary theorems. We first prove part $(1)$ and $(2)$ and establish that under mild conditions on $\ke$ we can extract a finite dimensional proxy kernel $\kc$ by truncating the eigen-decomposition of $\ke$ and estimating a feature map with samples. We leverage a kernel PCA result by \citet{zwald2006convergence} to construct $\kc$. 
\begin{theorem}\citep[Adapted from Theorem 4 in][]{zwald2006convergence}
\label{t:zwald}
If $m, p, S_m, \hat{S}_m, \delta_m, B_m$ and $\alpha$ are defined as in Theorem \ref{thm:proxy_properties} then with probability $1-\exp(-\alpha)$ we have
\begin{equation}
    \lVert P_{S_m} - P_{\hat{S}_m} \rVert_{F} \leq \frac{B_m}{\sqrt{p(m)}}.
\end{equation}
In particular,
\begin{equation*}
    \hat{S}_m \subset \left\{ g + h, g \in S_m, h \in S_m^{\perp}, \lVert h \rVert_\mathcal{H} \leq \frac{2B_m}{\sqrt{p}} \lVert g \rVert_\mathcal{H} \right\}.
\end{equation*}
\end{theorem}
Now using this theorem we prove Part $(1)$ of Theorem \ref{thm:proxy_properties}.
\begin{lemma} \label{lem:part1control}
With probability $1-e^{-\alpha}$ we have,
\begin{equation*}
    \lvert \ke(x,y) - \kc(x,y) \rvert \leq \frac{\epsilon}{4} + \sqrt{\epsilon}w + w^2 , \qquad \qquad  \forall \ x,y \in \mathcal{A}.
\end{equation*}
\end{lemma}
\begin{Proof}
First we show this holds for $x=y$. 
\begin{align*}
    \lVert \Phi(x) - P_{\hat{S}_m} (\Phi(x)) \rVert_{\mathcal{H}} &\overset{(i)}{\leq} \lVert \Phi(x) - P_{S_m} (\Phi(x)) \rVert_{\mathcal{H}} + \lVert P_{S_m} (\Phi(x)) - P_{\hat{S}_m} (\Phi(x)) \rVert_{\mathcal{H}} \\
    &\overset{(ii)}{\leq} \frac{\sqrt{\epsilon}}{2} + \lVert \Phi(x) \rVert_{\mathcal{H}} \lVert P_{S_m} - P_{\hat{S}_m} \rVert_{op}  \overset{(iii)}{\leq} \frac{\sqrt{\epsilon}}{2} + \mathcal{G}\frac{B_m}{\sqrt{p(m)}} \overset{(iv)}{\leq} \frac{\sqrt{\epsilon}}{2} + w,
\end{align*}
where $(i)$ follows by triangle inequality, $(ii)$ is by the fact that $P_{S_m}(\Phi(x))$ is an $\epsilon/4$ approximation of $\ke$, $(iii)$ follows by Theorem \ref{t:zwald} and $(iv)$ is by the choice of $p(m)$. Therefore with probability at least $1-e^{\alpha}$ for all $x \in \mathcal{A}$
\begin{equation*}
    \lvert \ke(x,x) - \kc(x,x)\rvert \leq  \frac{\epsilon}{4} + \sqrt{\epsilon}w + w^2.
\end{equation*}
Now we prove the statement for general $x,y \in \mathcal{A}$. We write $\Phi(x)  = \Phi_m(x) + h_x$ and $\Phi(y) = \Phi_m(y) + h_y$. The above calculation implies that $\lVert h_x \rVert_{\mathcal{H}} \leq  \frac{\sqrt{\epsilon}}{2} + w$ and $\lVert h_y \rVert_{\mathcal{H}} \leq \frac{\sqrt{\epsilon}}{2} + w$. We now expand $\ke(x,y)$ to get
\begin{equation*}
\langle \Phi(x), \Phi(y) \rangle_{\mathcal{H}} = \langle \Phi_m(x), \Phi_m(y) \rangle_{\mathcal{H}} + \langle h_x, \Phi_m(y) \rangle_{\mathcal{H}} + \langle \Phi_m(x), h_y \rangle_{\mathcal{H}} + \langle h_x ,  h_y \rangle_{\mathcal{H}}. 
\end{equation*}
Since $h_x$ and $h_y$ both live in $\hat{S}_m^{\perp}$:
\begin{equation*}
\langle \Phi(x), \Phi(y) \rangle_{\mathcal{H}} = \langle \Phi_m(x), \Phi_m(y) \rangle_{\mathcal{H}} + \langle h_x ,  h_y \rangle_{\mathcal{H}} .
\end{equation*}
Rearranging terms,
\begin{align*}
   \lvert \langle \Phi(x), \Phi(y) \rangle_{\mathcal{H}}  - \langle \Phi_m(x), \Phi_m(y) \rangle_{\mathcal{H}}\rvert &= \lvert\langle h_x ,  h_y \rangle_{\mathcal{H}} \rvert \leq \frac{\epsilon}{4} + \sqrt{\epsilon}w + w^2.
\end{align*}
This establishes the claim.
\end{Proof}
We now move on to the proof of Part $(2)$ in Theorem \ref{thm:proxy_properties}.
\begin{lemma}
If $p,B_m$ are chosen as stated in Theorem \ref{thm:proxy_properties} we have
\begin{equation}
\lvert \kc(x,y) - \kc^o(x,y)\rvert \leq w^2 \text{ } \qquad \qquad \forall x,y \in \mathcal{A}
\end{equation}
\end{lemma}
\begin{Proof}
The feature map for $\kc^o$ is $P_{S_m}(\Phi(x))$ for all $x \in \mathcal{A}$ while $P_{\hat{S}_m}(\Phi(x))$ is the feature map for $\kc$. We first show that for all $x \in \mathcal{A}$,
\begin{align*}
   \lVert P_{S_m}(\Phi(x)) - P_{\hat{S}_m}(\Phi(x)) \rVert_{\mathcal{H}} &\leq \lVert \Phi(x) \rVert_{\mathcal{H}} \lVert P_{S_m} - P_{\hat{S}_m} \rVert_{op} \leq \mathcal{G}\frac{B_m}{\sqrt{p(m)}}\leq w,
\end{align*}
where the second inequality follows by applying Theorem \ref{t:zwald} and the last inequality follows by the choice of $B_m$. A similar argument as the one used in the proof of Lemma \ref{lem:part1control} lets us then conclude that,
\begin{equation*}
|\kc(x,y) - \kc^o(x,y)| \leq w^2, \qquad \forall x,y \in \mathcal{A}.
\end{equation*}
\end{Proof}
We now proceed to prove part $(3)$ of Theorem \ref{thm:proxy_properties}. We will show that the Mercer operator eigenvalues of $\kc$ are close to Mercer operator eigenvalues of $\kc^{o}$. We first recall a useful result by \citet{mendelson2006singular}.
\begin{theorem}\citep[Adapted from Theorem 3.3 in][]{mendelson2006singular}\label{theorem:uniform_eigenvalues}
Let $\mathcal{K}$ be a kernel over $ \mathcal{A} \times \mathcal{A}$ such that $\sup_{x \in \mathcal{A}} \mathcal{K}(x,x) \leq \mathcal{G}^2$. Also let $\hat{\mu}_1 \geq \hat{\mu}_2 \geq \cdots \geq \hat{\mu}_N$ be the eigenvalues of the Gram matrix $(\ke(x_i, x_j)/N)_{i,j = 1}^N$ for $\{x_i\}_{i=1}^{N} \sim \mathbb{P}$. Then there exists a universal constant $c$ such that for every $t > 0$
\begin{equation}
\mathbb{P}\left[\sup_{i \in {1,\ldots,N}} \lvert \hat{\mu}_i - \mu_i \rvert \geq t\right ] \leq 2\exp\left( -\frac{ct}{\mathcal{G}^2}\sqrt{\frac{N}{\log(N)}} \right),
\end{equation}
where for $i > N$ we define $\hat{\mu}_i = 0$.
\end{theorem}
\begin{proposition}\label{prop:gramm_cov_equiv}
The top $m$ eigenvalues of the sample kernel covariance matrix equal that of the Gram matrix.
\end{proposition}
Recall that we established this proposition in Appendix \ref{app:kernelPCA} as Lemma \ref{lem:sameeigenvalues}. Further note that for any set of samples $x_1, \cdots, x_N \sim \mathbb{P}$, the Gram matrices of $\kc^o$ ($\mathbb{K}_m^o(N)$) and $\kc$ ($\mathbb{K}_m(N)$) are close in Frobenius norm as the matrices are close element-wise by part $(2)$ of Theorem \ref{thm:proxy_properties}.
\begin{equation*}
 \left \lVert \mathbb{K}_m^o(N) - \mathbb{K}_m(N) \right\rVert_F \leq w^2.
\end{equation*}
Let $\hat{\mu}^{(m,o)}_1 \geq \hat{\mu}^{(m,o)}_2 \geq \cdots \hat{\mu}^{(m,o)}_N$ and $\hat{\mu}^{(m)}_1 \geq \hat{\mu}^{(m)}_2 \geq \cdots \hat{\mu}^{(m)}_N$ be the eigenvalues of $\mathbb{K}_m^o(N)$ and $ \mathbb{K}_m(N)$ respectively. For both of these Gram matrices only the top $m$ out of $N$ eigenvalues will be nonzero, since both kernels are $m-$dimensional. By the Wielandt-Hoffman inequality \citep{hoffman1953variation} this implies that the \emph{ordered eigenvalues} are close,
\begin{equation*}
\sup_{i=1,\cdots N} |\hat{\mu}^{(m,o)}_i - \hat{\mu}^{(m)}_i| \leq w^2.
\end{equation*}
Theorem \ref{theorem:uniform_eigenvalues} and Proposition \ref{prop:gramm_cov_equiv} together imply the statement of Theorem \ref{theorem:uniform_eigenvalues} with the Gram matrix replaced by the sample covariance matrix holds. 
\begin{theorem}\label{theorem:mercer_sample_vs_chopped}
The Mercer operator eigenvalues $\mu^{(m)}_1 \geq \cdots \geq \mu^{(m)}_m$ and $\mu_1 \geq \cdots \geq \mu_m$ of $\kc$ and $\kc^o$ follow
\begin{equation}
\sup_{i=1, \cdots, m} |\mu^{(m)}_i - \mu_i| \leq w^2.
\end{equation}
\end{theorem}
\begin{Proof}
We will use the probabilistic method. By Theorem \ref{theorem:uniform_eigenvalues}, for every $t > 0$ there is $N(t) \in \mathbb{N}$ large enough such that probability of the event -- the eigenvalues of both sample Gram matrices $\mathbb{K}_m(N)$ and $\mathbb{K}_m^o(N)$ be uniformly close to the Mercer operator eigenvalues $\mu^{(m)}_1 \geq \cdots \geq \mu^{(m)}_m$ and $\mu_1 \geq \cdots \geq \mu_m$ -- is greater than zero.
By triangle inequality this implies that for all $t>0$
\begin{align*}
\sup_{i=1, \cdots, m} |\mu^{(m)}_i - \mu_i| &\leq \sup_{i_1} |\mu^{(m)}_{i_1} - \hat{\mu}^{(m)}_{i_1}      | + \sup_{i_2} |   \hat{\mu}^{(m)}_{i_2} - \hat{\mu}^{(m,0)}_{i_2}  | + \sup_{i_3}| \hat{\mu}^{(m,0)}_{i_3} -   \mu_{i_3} | \\& \leq  t + w^2 + t = w^2+2t.
\end{align*}
Taking the limit as $t \rightarrow 0$ yields the result.
\end{Proof}
\subsection{Bounds on the effective dimension $m$}\label{app:effectivedim}
In this section we establish bounds on the effective dimension $m$ under different eigenvalue decay assumptions.
\begin{proposition} \label{prop:mbounds}Let the conditions stated in Theorem \ref{thm:proxy_properties1} and Lemma \ref{lemma_chopped_kernel} hold. 
\begin{enumerate}
\item When the kernel $\ke$ has $(C,\beta)$-polynomial eigenvalue decay then 
\begin{align*}
m \ge \left[\frac{4C\mathcal{B}^2}{(\beta-1)\epsilon} \right]^{1/\beta-1},
\end{align*}
suffices for $\kc^o$ to be an $\epsilon/4$-approximation of $\ke$ and therefore for $\kc$ to be an $\epsilon$-approximation of $\ke$. 
\item When the kernel $\ke$ has $(C,\beta)$-exponential eigenvalue decay then 
\begin{align*}
m \ge \frac{1}{\beta}\log\left(\frac{4C \mathcal{B}^2}{\beta\epsilon}\right),
\end{align*}
suffices for $\kc^o$ to be an $\epsilon/4$-approximation of $\ke$ and therefore for $\kc$ to be an $\epsilon$-approximation of $\ke$. 
\end{enumerate}
\end{proposition}
\begin{Proof}
We need to ensure that the assumption in Lemma \ref{lemma_chopped_kernel} holds. That is,
\begin{align*}
\sum_{j=m+1}^{\infty} \mu_j \le \frac{\epsilon}{4\mathcal{B}^2}.
\end{align*}
We will prove the bound assuming a $(C,\beta)$-polynomial eigenvalue decay, the calculation is similar when we have exponential eigenvalue decay. Note that,
\begin{align*}
\sum_{j=m+1}^{\infty} \mu_j & \le \sum_{j=m+1}^{\infty} C j^{-\beta} \le \int_{m}^{\infty} C x^{-\beta} dx  = \frac{C}{\beta-1}\frac{1}{m^{\beta-1}}.
\end{align*}
We demand that,
\begin{align*}
\frac{C}{\beta-1}\frac{1}{m^{\beta-1}} &\le \frac{\epsilon}{4\mathcal{B}^2},
\end{align*}
rearranging terms yields the desired claim.
\end{Proof}
\section{Properties of the Covariance matrix -- $\Sigma_{m}^{(t)}$}
\label{app:covariancesample}
We borrow the notation from Section \ref{sec:constrproxy}. In this section we let $\mu_m$ be the smallest nonzero eigenvalue of $\mathbb{E}_{x \sim \nu}\left[ \Phi_m(x)\Phi_m(x)^\top \right]$ where $\nu$ is the exploration distribution over $\mathcal{A}$. 

\begin{lemma}\label{l:eigen_lowerbound_sigma_t}
Let $\mu_m^{(t)}$ be the $m$-th (smallest) eigenvalue of $\Sigma_m^{(t)}$. Then we have
\begin{equation*}
\mu_m^{(t)} \geq \gamma \mu_m.
\end{equation*}
\end{lemma}
\begin{Proof}
Recall that in each step we set $p_t = (1-\gamma) q_t + \gamma \nu$.  Let $v \in \mathcal{H}$ be a vector with norm 1. 
\begin{equation*}
v^\top \Sigma_m^{(t)} v = (1-\gamma)\cdot v^\top \mathbb{E}_{x \sim q_t}\left[    \Phi_m(x) \Phi_m(x)^\top\right] v+ \gamma \cdot v^\top \mathbb{E}_{x \sim \nu}\left[ \Phi_m(x) \Phi_m(x)^\top  \right]v.
\end{equation*}
Since both summands on the RHS are nonnegative, this quantity at least achieves a value of $\gamma \cdot v^\top \mathbb{E}_{x \sim \nu}\left[ \Phi_m(x) \Phi_m(x)^\top  \right]v \geq \gamma \mu_m$.
\end{Proof}
Observe that by our discussion in Appendix \ref{app:Johntheorem}, the minimum eigenvalue when the distribution is $\nu_{\mathcal{J}}$ (John's distribution) over $\Phi_m(\mathcal{A})$, then $\mu_m = 1/m$. That is, if $\nu_{\mathcal{J}}^{\mathcal{A}}$ is the exploration distribution over $\mathcal{A}$ then $\mu_m = 1/m$.
\begin{proposition}\label{l:johneigenvaluebound} If $\nu_{\mathcal{J}}^{\mathcal{A}}$ is the exploration distribution then we have
\begin{align*}
\mu_m^{(t)} \ge \frac{\gamma}{m}.
\end{align*}
\end{proposition}
\subsection{Finite Sample Analysis}
Next we analyze the sample complexity of the operation of building the second moment matrix in Algorithm \ref{a:expweightskernels} using samples. Let $\hat{\Sigma}_{m}^{(t)}$ be the second moment matrix estimate built by using $x_1, \cdots, x_r$ drawn i.i.d. from $p_t$.  
\begin{equation*}
\hat{\Sigma}_{m}^{(t)} = \frac{1}{r}\sum_{i=1}^r \Phi_m(x_i) \Phi_m(x_i)^\top.
\end{equation*}
We will show how to chose $r$ appropriately to preserve the validity of the regret bound when we use $\hat{\Sigma}_{m}^{(t)}$ (built using finite samples) instead of $\Sigma_{m}^{(t)}$. First we present some observations.
\begin{remark}[Covariance eigenvalues are Mercer's eigenvalues]
The eigenvalues $\mu_1^{(t)} \geq \cdots \geq \mu_m^{(t)}$ of $\mathbb{E}_{x\sim p_t}\left[ \Phi_m(x) \Phi_m(x)^\top \right]$ are exactly Mercer operator eigenvalues for $\kc$ under $p_t$. 
\end{remark}
\begin{remark}[Sample covariance and Gram matrix have the same eigenvalues]
Assume $r \geq m$. Let $x_1, \cdots, x_r \sim p_t$. The eigenvalues of the sample covariance $\hat{\Sigma}_{m}^{(t)}$ coincide with the top $m$ eigenvalues of the Gram matrix $\mathbb{K}_{m,p}^{(t)} = \left( \kc(x_i,x_j) \right)_{i,j = 1}^p$.
\end{remark}
We formalize the above remark in Lemma \ref{l:GtoS}. We will use an auxiliary lemma by \citet{zwald2006convergence} which we present here for completeness.
\begin{lemma}\citep[Lemma 1 in][]{zwald2006convergence}\label{lemma:covariance_operator_concentration}
Let $\mathcal{K}'$ be a kernel over $ \mathcal{X} \times \mathcal{X}$ such that $\sup_{x \in \mathcal{X}} \mathcal{K}'(x,x) \leq \mathcal{G}'$. Let $\Sigma'$ be the covariance of $\Phi'(x)$, $x \sim \mathbb{P}$. If $\hat{\Sigma}'_r$ is the sample covariance built by using $r$ samples $x_1, \cdots, x_r \sim \mathbb{P}$, with probability $1-\exp(-\delta)$:
\begin{equation*}
\lVert \Sigma' - \hat{\Sigma}'_p \rVert_{op} \leq \frac{2\mathcal{G}'}{\sqrt{r}} \left( 1 + \sqrt{\frac{\delta}{2}}\right).
\end{equation*}
\end{lemma}
The following lemma will allow us to derive an operator norm bound between the inverse matrices $\left( \Sigma_{m}^{(t)}\right)
^{-1}$ and $\left(\hat{\Sigma}_{m}^{(t)}\right)^{-1}$ from an operator norm bound between the matrices $\Sigma_{m}^{(t)}$ and $\hat{\Sigma}_{m}^{(t)}$.
\begin{lemma}\label{lemma:operator_between_inverses}
If $\lVert A - B \rVert_{op} \leq s$, then $\lVert A^{-1} - B^{-1}\rVert_{op} \leq \frac{s}{\lambda_{min}(A) \lambda_{min}(B)} $, where $\lambda_{min}(A)$ and $\lambda_{min}(B)$ denote the minimum eigenvalues of $A$ and $B$ respectively.
\end{lemma}
\begin{Proof}
The following equality holds:
\begin{equation*}
A^{-1} - B^{-1} = A^{-1}( B- A ) B^{-1}.
\end{equation*}
Applying Cauchy-Schwartz for spectral norms yields the desired result.
\end{Proof}

We are now ready to show that given enough samples $r$, the operator norm between the inverse covariance and the inverse sample covariance is small.

\begin{lemma} \label{l:mainsamplebound}
Let $g: \mathbb{R}^+_{(1, \infty)} \rightarrow \mathbb{R}$ be defined as $g(x)$ is the value such that $\frac{g(x)}{\log(g(x))} = x$. 
If the number of samples
\begin{align*}
r \geq \max\left( g\left( \left[ \frac{(\ln(2) + \zeta)2\mathcal{G}}{c \gamma \mu_m}   \right]^2\right),  \left( \frac{4\mathcal{G}(1+\sqrt{\frac{\zeta}{2}})}{(\gamma\mu_m)^2\epsilon_1}   \right)^2   \right),
\end{align*} where $c$ is the same constant as in Theorem \ref{theorem:uniform_eigenvalues}, then with probability $1-2e^{-\zeta}$:
\begin{equation*}
\left \lVert \left( \Sigma_{m}^{(t)}\right)^{-1} - \left( \hat{\Sigma}_{m}^{(t)}\right)^{-1} \right\rVert_{op} \leq \epsilon_1.
\end{equation*}
\end{lemma}
\begin{Proof} We start by showing that if $r$ follows the requirements stated in the lemma above, then the minimum eigenvalue of $\hat{\Sigma}_{m}^{(t)}$ is lower bounded by $\frac{\gamma\mu_m}{2}$ with probability $1-\exp(-\zeta)$. We invoke Theorem \ref{theorem:uniform_eigenvalues} to prove this. Let us denote by $\mu_1^{(t)} \geq \cdots \geq \mu_m^{(t)}$ and $\hat{\mu}_1^{(t)} \geq \cdots \geq \hat{\mu}_m^{(t)}$ the eigenvalues of $\Sigma_{m}^{(t)}$ and $\hat{\Sigma}_{m}^{(t)}$ respectively.

We want to ensure that the probability of $\sup_{i} | \mu_i^{(t)} - \hat{\mu}_i^{(t)}| \geq \frac{\gamma\mu_m}{2}$ be less than $e^{-\zeta}$. Again by invoking Theorem \ref{theorem:uniform_eigenvalues}, this is true if $exp(-\zeta) \le 2\exp\left( -\frac{c\gamma\mu_m}{2\mathcal{G}}\sqrt{\frac{r}{\log(r}}\right)$. This yields the condition,
\begin{equation*}
\frac{r}{\log(r)} \ge \left[\frac{(\ln(2)+\zeta) 2\mathcal{G}}{c\gamma \mu_m}   \right]^2.
\end{equation*}
This together with triangle inequality (as $\mu_m^{(t)} \ge \mu_m \ge \gamma \mu_m$) ensures that if $r \geq g\left( \left[ \frac{(\ln(2) + \zeta)2\mathcal{G}}{c \gamma \mu_m}   \right]^2\right)$, then with probability $1-\exp(-\zeta)$,
\begin{align*}
\hat{\mu}_m \geq \frac{\gamma \mu_m}{2}.
\end{align*}
Setting $A = \Sigma_{m}^{(t)}$ and $B = \hat{\Sigma}_{m}^{(t)}$ and invoking the concentration inequality Lemma \ref{lemma:covariance_operator_concentration}, we have that 
\begin{align*}
\left\lVert \Sigma_{m}^{(t)} - \hat{\Sigma}_{m}^{(t)}\right\rVert_{op} \leq \frac{(\gamma\lambda_m)^2\epsilon_1}{2},
\end{align*} 
with probability $1-\exp(-\zeta)$ as choose $r$ to satisfy
\begin{equation*}
\frac{2\mathcal{G}}{2\sqrt{r}} \left( 1+ \sqrt{\frac{\zeta}{2}} \right) = \frac{(\gamma\mu_m)^2\epsilon_1}{2}.
\end{equation*}
As the matrices $A$ and $B$ are close with high probability, Lemma \ref{lemma:operator_between_inverses} proves that the inverses are also close,
\begin{align*}
\left \lVert \left( \Sigma_{m}{(t)}\right)^{-1} - \left(\hat{\Sigma}_{m}^{(t)}\right)^{-1}\right\rVert_{op} \leq \epsilon_1.
\end{align*}
with the same probability. By union bound as long as $r \geq \max\left( g\left( \left[ \frac{(\ln(2) + \zeta)2\mathcal{G}}{c \gamma \lambda_m}   \right]^2\right),  \left( \frac{4\mathcal{G}(1+\sqrt{\frac{\zeta}{2}})}{(\gamma\lambda_m)^2\epsilon_1}   \right)^2   \right)$ the stated claim holds with probability $1-2\exp(-\zeta)$.
\end{Proof}

\subsubsection{Auxiliary Lemmas}
Let us denote the pseudo-inverse of a symmetric matrix $A$ by $A^{\dagger}$. We now prove Lemma \ref{l:GtoS} that formalizes the connection between the eigenvalues the Gram matrix and sample covariance matrix.
\begin{lemma} \label{l:GtoS}
For any $x,y \in \mathcal{A}$:
\begin{equation*}
\Phi_m(x)^\top \left( \hat{\Sigma}_{m}^{(t)} \right)^{-1} \Phi_m(y) = A_x^\top \left( \mathbb{K}_{m,p}^{(t)}\right)^{2\dagger} A_y^\top,
\end{equation*}
where $A_x = \left(\kc(x, x_1), \cdots, \kc(x,x_p) \right)^\top$ and $A_y = \left(\kc(y, x_1), \cdots, \kc(y,x_p) \right)^\top$.
\end{lemma}

\begin{Proof}
The claim can be verified by a singular value decomposition of both sides.
\end{Proof}

Given Lemma \ref{l:mainsamplebound} we also prove a bound on the distance between the estimates of adversarial actions generated in Algorithm \ref{a:expweightskernels}. Define $\tilde{w}^{(2)}_t:= \left(\hat{\Sigma}_{m}^{(t)}\right)^{-1} \Phi_m(a_t)\ke(a_t,w_t)$ and let $\hat{w}_t := \tilde{w}^{(1)}_t= \left(\Sigma_{m}^{(t)}\right)^{-1} \Phi_m(a_t)\ke(a_t,w_t)$.
\begin{corollary} We have that
\begin{equation*}
\lVert  \tilde{w}^{(2)}_t - \tilde{w}^{(1)}_t \rVert_{\mathcal{H}} \leq \epsilon_1 \mathcal{G}.
\end{equation*}
\end{corollary}
In other words, the bias resulting from using the sample covariance instead of the true covariance is of order $\epsilon_1$ as long as we take enough samples $p$ at each time step. We can drive $\epsilon_1$ to be as low as we like by choosing enough samples and hence this bias does not determine the rate in the regret bounds in Theorem \ref{thm:mainregretbound}.
\section{Full Information Regret Bounds} \label{app:fullinfo}
\subsection{Exponential Weights Regret Bound}
 \label{app:regretexp}
In this section we prove a regret bound for exponential weights and present a proof of Theorem \ref{t:expweightsregret}. The analysis of the regret is similar to the analysis of exponential weights for linear losses \citep[see for example a review in][]{peter}. In the proof below we denote the filtration at the end of round $t$ by $\mathcal{F}_t$, that is, it conditions on the past actions of the player and the adversary ($a_{t-1},w_{t-1},\ldots,a_1,w_1$).
\begin{Proof}[Proof of Theorem \ref{t:expweightsregret}]
By the tower property and by the definition of the regret we can write the cumulative loss as,
\begin{align*}
\mathbb{E}\left[\sum_{t=1}^{n} \langle \Phi(a_t), w_t \rangle \right] & = \mathbb{E}\left[ \sum_{t=1}^n \mathbb{E}_{a_t \sim p_t}\left[ \langle \Phi(a_t), w_t \rangle \Bigg\rvert  \mathcal{F}_{t-1}\right]\right] \\
& = \mathbb{E}\left[ \sum_{t=1}^n  \left[\int_{\mathcal{A}} p_t(a) \langle \Phi(a), w_t \rangle da \Bigg\rvert  \mathcal{F}_{t-1}\right]\right].
\end{align*}

Observe that our choice of $\eta$ implies that $\eta \langle \Phi(a), w_t \rangle > -1$. By invoking Hoeffding's inequality (stated as Lemma \ref{hff}) we get
\begin{align*}
& \mathbb{E}\left[\sum_{t=1}^{n} \left[\int_{\mathcal{A}}  p_t(a) \langle \Phi(a), w_t \rangle da \Bigg\rvert \mathcal{F}_{t-1} \right] \right] \\
& \le -\frac{1}{\eta} \underbrace{\mathbb{E}\left[\sum_{t=1}^{n} \log\left( \mathbb{E}_{a \sim p_t}\left[ \exp\left(-\eta \langle \Phi(a), w_t \rangle\right)\lvert \mathcal{F}_{t-1}\right]\right)\right]}_{=:\Gamma} + (e-2) \eta \mathbb{E}\left[\sum_{t=1}^n \int_{ \mathcal{A}} \left[p_t(a) \langle \Phi(a), w_t \rangle^2 da\lvert \mathcal{F}_{t-1}\right] \right] \\
& \overset{(i)}{\le} -\frac{\Gamma}{\eta} + (e-2)\eta \mathcal{G}^4 n,
\end{align*}
where $(i)$ follows by Cauchy-Schwartz and the bound on the adversarial and player actions. Next we bound $\Gamma$ using Lemma \ref{l:gammabound}. Substituting this bound into the expression above we get
\begin{align*}
\mathbb{E}\left[\sum_{t=1}^{n} \langle \Phi(a_t), w_t \rangle \right] 
 & \le \mathbb{E}\left[\sum_{t=1}^{n} \langle \Phi(a^*),w_t\rangle  \right] + \frac{\log(vol(\mathcal{A}))}{\eta} + (e-2)\eta \mathcal{G}^4 \cdot n.
\end{align*}
Rearranging terms we have the regret is bounded by
\begin{align*}
\mathcal{R}_n & \le  (e-2) \mathcal{G}^4 \eta  n + \frac{\log(vol(\mathcal{A}))}{\eta}.
\end{align*}
The choice of $\eta = \sqrt{\log(vol(\mathcal{A}))}\Big/\sqrt{(e-2)}\mathcal{G}^2n^{1/2}$, optimally trades of the two terms to establish a regret bound of $\mathcal{O}(n^{1/2})$.
\end{Proof}
Next we provide a proof of the bound on $\Gamma$ used above.
\begin{lemma}\label{l:gammabound} Assume that $p_1(\cdot)$ is chosen as the uniform distribution in Algorithm \ref{a:expweightskernels}. Also let $\Gamma$ be defined as follows
\begin{align*}
\Gamma = \mathbb{E}\left[\sum_{t=1}^n \log\left( \mathbb{E}_{a\sim p_t} \left[\exp\left( -\eta \langle \Phi(a),w_t\rangle_{\mathcal{H}}\right)\Big\lvert \mathcal{F}_{t-1}\right]\right)\right].
\end{align*}
Then we have that,
\begin{align*}
\Gamma & \ge -\eta\mathbb{E}\left[ \sum_{i=1}^{n}\langle \Phi(a^*), w_i \rangle_{\mathcal{H}} \right]  -\log\left( vol(\mathcal{A})\right),
\end{align*}
where $a^*$ is the optimal action in hindsight in the definition of regret and $vol(\mathcal{A})$ is the volume of the set $\mathcal{A}$.
\end{lemma}
\begin{Proof}Expanding $\Gamma$ using the definition of $p_t$ we have that,
\begin{align*}
\Gamma & 
 \overset{(i)}{=} \mathbb{E}\left[\sum_{t=1}^{n} \log\left\{ \frac{\int_{\mathcal{A}}\exp\left( -\eta \sum_{i=1}^{t}\langle \Phi(a), w_i \rangle_{\mathcal{H}}\right)da}{ \int_{\mathcal{A}} \exp\left( -\eta \sum_{i=1}^{t-1}\langle \Phi(a), w_i \rangle_{\mathcal{H}}\right)da} \right\}\right]\\
 & \overset{(ii)}{=} \mathbb{E}\left[\log\left(\int_{\mathcal{A}}\exp\left( -\eta \sum_{i=1}^{n}\langle \Phi(a), w_i \rangle_{\mathcal{H}}\right) da\right) \right] -\log\left( vol( \mathcal{A} )\right), 
\end{align*}
where $(i)$ follows by the definition of $p_t(a)$ and $(ii)$ is by expanding the sum and canceling the terms in a telescoping series. The $\log(vol( \mathcal{A}))$ term is because we start off with a uniform distribution over all elements. Lastly observe that by optimality of $a^*$ we have that,
\begin{align*}
\mathbb{E}\left[\log\left(\int_{\mathcal{A}}\exp\left( -\eta \sum_{i=1}^{n}\langle \Phi(a), w_i \rangle\right) da\right) \right] & \ge -\eta \mathbb{E}\left[\sum_{i=1}^{n}\langle \Phi(a^*), w_i \rangle  \right].
\end{align*}
Plugging this into the above expression establishes the desired bound on $\Gamma$.
\end{Proof}
We now present a proof of Lemma \ref{lem:quadraefficient} that guarantees that it is possible to sample efficiently from the exponential weights distribution when the losses are quadratics.
\begin{Proof}[Proof of Lemma \ref{lem:quadraefficient}]
Let $v_1, \cdots, v_d$ an orthonormal basis of eigenvectors of $B$ with eigenvalues $\lambda_1, \cdots, \lambda_d$ possibly negative. We express $b$ using the basis $\{v_i\}_{i=1}^d$ as $b = \sum_{i=1}^d \gamma_i v_i$. Also let $a = \sum_{i=1}^d \alpha_i v_i$. By the definition of the set $\mathcal{A}$ we have $\sum_{i=1}^d \alpha_i^2 \le 1$. The distribution $q(\cdot)$ can be thus expressed as
\begin{align*}
q(a) \propto \exp\left( \sum_{i=1}^d (\lambda_i \alpha_i^2 + \gamma_i \alpha_i) \right).
\end{align*}
Completing the squares (whenever $\lambda_i \neq 0$),
\begin{align*}
q(a) \propto \exp\left\{ \sum_{i=1}^d \lambda_i \left( \alpha_i^2 + \frac{\gamma_i \alpha_i}{\lambda_i} + \left(\frac{\gamma_i}{2\lambda_i}\right)^2\right) \right\}.
\end{align*} 
Let us re-parametrize this distribution by setting $\beta_i = (\alpha_i + \frac{\gamma_i}{2\lambda_i})^2$. The inverse mapping is $\alpha_i=\sqrt{\beta_i} - \frac{\gamma_i}{2\lambda_i}$. To sample from $q(\cdot)$ it is enough to produce a sample from a surrogate distribution $\beta \sim t(\beta)$ and turn them into a sample of $q$ where,
\begin{align*}
t(\beta) & \propto \exp\left( \sum_{i=1}^d \lambda_i \beta_i \right),\\
s.t. \qquad 0 & \leq \beta_i, \\
\sum_{i=1}^d \left(\sqrt{\beta_i} - \frac{\gamma_i}{2\lambda_i} \right)^2 & \leq 1.
\end{align*}
Let $\{\epsilon_i\}_{i = 1}^{d}$ be independent Bernoulli $\{-1, 1\}$ variables, then $ a = \sum_{i=1}^d  \epsilon_i (\sqrt{\beta_i}-\frac{\gamma_i}{2\lambda_i})v_i$ is a sample from $q$. Note that the distribution $t(\beta)$ is log-concave. We now show that the constraint set $\mathcal{C}$ is convex, where $\mathcal{C}= \{\beta | \beta_i \geq 0, \sum_{i=1}^d (\sqrt{\beta_i} - \frac{\gamma_i}{2\lambda_i} )^2 \leq 1\}$. 

Let $\hat{\beta}$ and $\tilde{\beta}$ be two distinct points in $\mathcal{C}$. We show that for any $\eta \in [0,1]$ the point $\eta \hat{\beta} + (1-\eta) \tilde{\beta} \in \mathcal{C}$. The non-negativity constraint is clearly satisfied $(\eta \hat{\beta} + (1-\eta) \tilde{\beta} )_i\geq 0, \forall i$. The second constraint can be rewritten as
\begin{align}
\sum_{i=1}^d \hat{\beta}_i - \frac{\gamma_i\sqrt{\hat{\beta}_i}}{\lambda_i} + \left(\frac{\gamma_i}{2\lambda_i}\right)^2 \leq 1 \label{hat_beta} \\
\sum_{i=1}^d \tilde{\beta}_i - \frac{\gamma_i\sqrt{\tilde{\beta}_i}}{\lambda_i} + \left(\frac{\gamma_i}{2\lambda_i}\right)^2 \leq 1 \label{tilde_beta}.
\end{align}
These equations imply that,
\begin{align*}
&\eta\left[\sum_{i=1}^d \hat{\beta}_i - \frac{\gamma_i\sqrt{\hat{\beta}_i}}{\lambda_i} + \left(\frac{\gamma_i}{2\lambda_i}\right)^2\right] 
+ (1-\eta)\left[ \sum_{i=1}^d \tilde{\beta}_i - \frac{\gamma_i\sqrt{\tilde{\beta}_i}}{\lambda_i} + \left(\frac{\gamma_i}{2\lambda_i}\right)^2 \right] \le 1.
\end{align*}
By concavity of the square root function we have
\begin{align*}
\sum_{i=1}^d \eta\frac{\gamma_i\sqrt{\hat{\beta}_i}}{\lambda_i} + (1-\eta)\frac{\gamma_i\sqrt{\tilde{\beta}_i}}{\lambda_i} \leq \sum_{i=1}^d \frac{\gamma_i \sqrt{  \eta \hat{\beta_i} + (1-\eta) \tilde{\beta_i}} }{\lambda_i},
\end{align*} 
these two observations readily imply that $\eta \hat{\beta} + (1-\eta) \tilde{\beta}$ satisfies the constraint of $\mathcal{C}$ thus implying convexity of $\mathcal{C}$. We can thus use Hit-and-Run \citep{lovasz2007geometry} to sample from $t(\beta)$ in $\tilde{O}(d^4)$ steps and convert to samples from $q(\cdot)$ using the method described above. In case some eigenvalues are zero, say without loss of generality $\lambda_1, \cdots, \lambda_R$. Then set $\beta_i = \alpha_i^2$ for $i \in \{ R+1,\ldots, d\}$ and sample from the distribution,
\begin{align*}
t(\beta) &\propto \exp\left(\sum_{i=1}^R \gamma_i \alpha_i +  \sum_{i=R+1}^d \lambda_i \beta_i\right),\\
s.t. \qquad 0 &\leq \beta_i,\\
\sum_{i=1}^R \alpha_i^2 & + \sum_{i=R+1}^d \left(\sqrt{\beta_i} - \frac{\gamma_i}{2\lambda_i} \right)^2  \leq 1.
\end{align*}
The analysis follows as before for this case as well.
\end{Proof}
\subsection{Conditional Gradient Method Analysis}
\label{frankwolferegretappendix}
\label{s:cdmregret}

The regret bound analysis for Algorithm \ref{a:CG_algo}, conditional gradient method over RKHSs follows by similar arguments to the analysis of the standard online conditional gradient descent \citep[see for example review in][]{hazan2016introduction}. To prove this we first prove the regret bound of a different algorithm -- follow the regularized leader. 
\subsection{Follow the Regularized Leader}
\begin{algorithm}[t]

\SetKwInOut{Input}{Input}
    \SetKwInOut{Output}{Output}
    \Input{Set $\mathcal{A}$, number of rounds $n$, initial action $a_{1} \in \mathcal{A}$, inner product $\langle \cdot,\cdot \rangle_{\mathcal{H}}$, learning rate $\eta >0 $. }
    
    Let $X_1 = \argmin_{X \in conv(\Phi(\mathcal{A}))} \frac{1}{\eta}\langle X,X \rangle$
    
    choose $\mathcal{D}_{1}$ such that $\mathbb{E}_{x \sim \mathcal{D}_1}[\Phi(x)] = X_1$
    
    \For {$t=1,2,3 \ldots, n$} 
      {
         choose $a_t \sim \mathcal{D}_t$
         
         observe $\langle \Phi(a_t),w_t \rangle_{\mathcal{H}}$
         
         update  $X_{t+1}  = \argmin_{X \in conv(\Phi(\mathcal{A}))} \eta \sum_{s = 1}^{t} \langle w_s , X \rangle_{\mathcal{H}} + \langle X, X \rangle_{\mathcal{H}} $
         
        choose $\mathcal{D}_{t+1}$ s.t. $\mathbb{E}_{x \sim \mathcal{D}_{t+1}}[\Phi(x)] = X_{t+1} $
   }     
\caption{Follow the Regularized leader (FTRL) \label{FTRL}} 
\end{algorithm}
We present a version of follow the regularized leader \citep{shalev2007primal} (FTRL, Algorithm \ref{FTRL}) adapted to our setup. Note that this algorithm is not tractable in general as at each step we are required to perform an optimization problem over the convex hull of $\Phi(\mathcal{A})$. However, we provide a regret bound that we will use in our regret bound analysis for the conditional gradient method. Let us define $w_0 = X_1/\eta$. We first establish the following lemma. 
\begin{lemma}[No regret strategy] \label{noregret} For any $u \in \mathcal{A}$
\begin{align*}
\sum_{t=0}^n \langle X_t , \Phi(u) \rangle_{\mathcal{H}} \ge \sum_{t=0}^n \langle X_t , X_{t+1} \rangle_{\mathcal{H}}.
\end{align*}
\end{lemma}
This is the crucial lemma needed to prove regret bounds for FTRL algorithms and its  proof follows from standard arguments \citep[see for example Lemma 5.3][]{hazan2016introduction}. 
\begin{definition} \label{gt}Define a function $g_{t}(\cdot): \mathbb{R}^D \mapsto R$ as,
\begin{align*}
g_{t}(X) \triangleq  \left[ \eta \sum_{s=1}^{t} \langle w_s, X \rangle_{\mathcal{H}} + \langle X,X \rangle_{\mathcal{H}} \right].
\end{align*}
\end{definition}
\begin{definition} \label{breg}
Define the Bregman divergence as,
\begin{align*}
B_{R}(x||y) \triangleq R(x) - R(y) - \langle \nabla R(y),(x-y) \rangle_{\mathcal{H}}.
\end{align*}
\end{definition}
Given these two definitions we now establish a lemma that will be used used to control the regret of FTRL.
\begin{lemma} \label{lossupperlemma}
For any $t \in \{1,2,\ldots, n \}$ we have the upper bound,
\begin{align*}
\langle w_t,X_t - X_{t+1} \rangle_{\mathcal{H}} & \le 2\eta \lVert w_t \rVert^2_{\mathcal{H}}.
\end{align*}
\end{lemma}
\begin{Proof}
By the definition of Bregman divergence we have,
\begin{align*}
g_{t}(X_t) & = g_{t}(X_{t+1}) + \langle X_t - X_{t+1}, \nabla g_{t}(X_{t+1}) \rangle_{\mathcal{H}} + B_{g_{t}}(X_t || X_{t+1})\\
& \ge g_{t}(X_{t+1}) + B_{g_{t}}(X_{t} || X_{t+1}),
\end{align*}
where the inequality is because $X_{t+1}$ is the minimizer of $g_{t}(\cdot)$ over $ conv(\Phi(\mathcal{X}))$. After rearranging terms we are left with an upper bound on the Bregman divergence,
\begin{align}
B_{g_{t}}(X_{t} || X_{t+1}) & \le g_{t}(X_{t}) - g_{t}(X_{t+1}) \nonumber\\
& = \big( g_{t-1}(X_{t}) - g_{t-1}(X_{t+1})\big) + \eta \langle w_t, X_{t} - X_{t+1} \rangle_{\mathcal{H}} \nonumber \\
& \le \eta \langle w_t, X_{t} - X_{t+1} \rangle_{\mathcal{H}}, \label{gequation}
\end{align}
where the last inequality follows because $X_{t-1}$ is the minimizer of the function $g_{t-1}(\cdot)$ over $conv(\Phi(\mathcal{X}))$. Observe that $B_{g_{t}}(X_{t}||X_{t+1}) = \frac{1}{2} \lVert X_{t} - X_{t+1}\rVert_{\mathcal{H}}^{2}$. Thus by the Cauchy-Schwartz inequality we have,
\begin{align*}
 \langle w_t, X_{t}- X_{t+1}) \rangle_{\mathcal{H}} & \le \lVert w_{t}(X_t) \rVert_{\mathcal{H}} \lVert X_{t} - X_{t+1} \rVert_{\mathcal{H}} \\
& = \lVert w_{t} \rVert_{\mathcal{H}} \sqrt{ 2 B_{g_{t}}(X_{t}||X_{t+1})}. 
\end{align*}
Substituting the upper bound from Equation \eqref{gequation} we get,
\begin{align*}
\langle w_t, X_{t}- X_{t+1}) \rangle_{\mathcal{H}} & \le \lVert w_{t} \rVert_{\mathcal{H}} \cdot \sqrt{2\eta \langle w_t, X_{t} - X_{t+1} \rangle_{\mathcal{H}}}.
\end{align*}
Rearranging terms establishes the result.
\end{Proof}
\begin{theorem} \label{FTRLregret}
Given a step size $\eta >0$, the regret suffered by Algorithm \ref{FTRL} after $n$ rounds is bounded by
\begin{align*}
\mathcal{R}_n & \le 2 n\eta \mathcal{G}^2 + \frac{2 \mathcal{G}^2}{\eta}.
\end{align*}
\end{theorem}
\begin{Proof}
By the definition of regret we have
\begin{align}
\mathcal{R}_n & =  \mathbb{E} \left[  \sum_{t=1}^n \langle \Phi(a_t),w_t\rangle_{\mathcal{H}} -\min_{a \in \mathcal{A}}\left[\sum_{t=1}^n  \langle \Phi(a_t),w_t \rangle_{\mathcal{H}}\right] \right] \nonumber\\
& \overset{(i)}{=} \mathbb{E}\left[\sum_{t=1}^n \mathbb{E}_{a_t \sim \mathcal{D}_t}\left[ \langle w_t,\Phi(a_t) - \Phi(a^*) \rangle_{\mathcal{H}} \Bigg \lvert \mathcal{F}_{t-1} \right] \right]\nonumber \overset{(ii)}{=} \mathbb{E} \left[ \sum_{t=1}^n  \langle w_t,X_t - \Phi(a^{*}) \rangle_{\mathcal{H}}   \right] \nonumber\\
&\overset{(iii)}{\le} \mathbb{E} \left[ \sum_{t=1}^{n}   \langle w_t, X_t - X_{t+1} \rangle_{\mathcal{H}} \right]  + \mathbb{E} \left[ \sum_{t=1}^{n}   \langle w_t, X_{t+1} - \Phi(a^*) \rangle_{\mathcal{H}} \right] + \frac{1}{\eta} (\langle X_1,X_1 \rangle_{\mathcal{H}} - (\langle X_0,X_0 \rangle_{\mathcal{H}})\nonumber\\
& \overset{(iv)}{\le} \mathbb{E} \left[ \sum_{t=1}^{n}   \langle w_t, X_t - X_{t+1} \rangle_{\mathcal{H}} \right] + \frac{2\mathcal{G}^2}{\eta}. \label{e:regretftrlstep}
\end{align}
The first equality follows as $a^*$ is the minimizer, $(ii)$ is by evaluating the expectation with respect to $\mathcal{D}_t$, $(iii)$ is an algebraic manipulation and finally $(iv)$ follows by invoking Lemma \ref{noregret} and using Cauchy-Schwartz to bound the last term. We need to control the first term in Equation \eqref{e:regretftrlstep} to get a regret bound. To control the first term we now invoke Lemma \ref{lossupperlemma}
\begin{align*}
\mathcal{R}_n & \le 2\eta\sum_{t=1}^n \lVert w_t \rVert_{\mathcal{H}}^2 + \frac{2\mathcal{G}^2}{\eta}  \le 2 n\eta \mathcal{G}^2 + \frac{2\mathcal{G}^2}{\eta}.
\end{align*}
This establishes the stated result.
\end{Proof}
\subsection{Regret Bound for Algorithm \ref{a:CG_algo}}
In deploying Algorithm \ref{a:CG_algo} we will at each round find distributions over the action space $\mathcal{A}$ as the player is only allowed play rank 1 actions in the Hilbert space at each round, while the action prescribed by the conditional gradient method might not be rank 1. Thus we find a distribution $\mathcal{D}_t$ such that,
\begin{align*}
\mathbb{E}_{a \sim \mathcal{D}_t} \Phi(a) = X_t,
\end{align*}
where $X_t$ is the action prescribed by Algorithm \ref{a:CG_algo}. We will strive to match the optimal action in expectation by choosing an appropriate distribution and get bounds on expected regret. For all $t \in \{1,2,\ldots,n \}$ let $X_t^{*}$ be defined as the iterates of the follow the regularized leader (Algorithm \ref{FTRL}) with the regularization  set to $R(X) = \lVert X - X_1 \rVert_{\mathcal{H}}^2$ and applied to the shifted loss function, $\langle w_t, X - (X_t^* - X_t)\rangle_{\mathcal{H}}$. Notice that,
\begin{align} \label{e:lipshitzactions}
\left\lvert \langle X,w_t \rangle_{\mathcal{H}} - \langle X-(X_t^*-X_t),w_t \rangle_{\mathcal{H}} \right\rvert & \le \lVert w_t \rVert_{\mathcal{H}} \lVert X_t^* - X_t \rVert_{\mathcal{H}} \le \mathcal{G} \lVert X_t^* - X_t \rVert_{\mathcal{H}}.
\end{align}
We are now ready to prove Theorem \ref{t:cgmregret}.
\begin{Proof}[Proof of Theorem \ref{t:cgmregret}] We denote the filtration up to round $t$ by $\mathcal{F}_{t-1}$, that is, we condition on all past player and adversary actions. Also let us denote the optimal action in hindsight by $a^*$. We begin by expanding the definition of regret to get,
\begin{align*}
\mathcal{R}_n & = \mathbb{E}\left[\sum_{t=1}^n \mathbb{E}_{a_t \sim \mathcal{D}_t} \left[\langle w_t,\Phi(a_t) \rangle_{\mathcal{H}} - \langle w_t,\Phi(a^*) \rangle_{\mathcal{H}} \ \Bigg \lvert \mathcal{F}_{t-1}\right] \right] \\
& \overset{(i)}{=} \mathbb{E} \left[\sum_{t=1}^n \langle w_t, X_t - \Phi(a^*) \rangle_{\mathcal{H}} \right]  = \mathbb{E} \left[\sum_{t=1}^n \langle w_t, X_t -X_t^* \rangle_{\mathcal{H}} \right] + \mathbb{E} \left[\sum_{t=1}^n \langle w_t, X_t^* -\Phi(a^*) \rangle_{\mathcal{H}} \right]\\ 
& \overset{(ii)}{\le} \mathbb{E} \left[\sum_{t=1}^n \langle w_t, X_t - X_t^* \rangle_{\mathcal{H}} \right] + 2 n \eta \mathcal{G}^2 + \frac{2\mathcal{G}^2}{\eta}  \overset{(iii)}{\le} \underbrace{\mathbb{E} \left[\sum_{t=1}^n \lVert w_t \rVert_{\mathcal{H}} \lVert X_t - X_t^* \rVert_{\mathcal{H}} \right] }_{=:\Xi}+ 2 n \eta \mathcal{G}^2 + \frac{2\mathcal{G}^2}{\eta},
\end{align*}
where $(i)$ follows by taking expectation with respect to $\mathcal{D}_t$, $(ii)$ follows by invoking Theorem \ref{FTRLregret} and $(iii)$ by Cauchy-Schwartz inequality. We finally need to control $\Xi$ to establish a bound on the regret.
\begin{align*}
\Xi & = \mathbb{E} \left[\sum_{t=1}^n \lVert w_t \rVert_{\mathcal{H}} \lVert X_t - X_t^* \rVert_{\mathcal{H}} \right]  \overset{(i)}{\le} \mathbb{E} \left[\sum_{t=1}^n \lVert w_t \rVert_{\mathcal{H}} \sqrt{F_t(X_t)- F_t(X_t^*)} \right] \overset{(ii)}{\le} 2\sum_{t=1}^n \mathcal{G}^2\sqrt{\gamma_t},
\end{align*}
here $(i)$ follows by the strong convexity of $F_t(\cdot)$ and $(ii)$ follows by the upper bound established in Lemma \ref{cgmhbound}. Plugging this into the bound for regret we have
\begin{align*}
\mathcal{R}_n & \le 2 \sum_{t=1}^n \mathcal{G}^2 \sqrt{\gamma_t} + 2n\eta \mathcal{G}^2 + \frac{2\mathcal{G}^2}{\eta} \overset{(i)}{\le} 4 \mathcal{G}^2n^{3/4} + 2n\eta \mathcal{G}^2 + \frac{2\mathcal{G}^2}{\eta},
\end{align*}
where $(i)$ follows by summing the series $1/t^{1/4}$ ($\sqrt{\gamma_t}$). The choice $\eta =1/n^{3/4}$ satisfies the conditions of Lemma \ref{cgmhbound} and we can plug in this choice to get,
\begin{align*}
\mathcal{R}_n & \le 4\mathcal{G}^2n^{3/4}+2\mathcal{G}^2n^{1/4}+ 2\mathcal{G}^2n^{3/4} \le 8\mathcal{G}^2n^{3/4}.
\end{align*}
This establishes the desired bound on the regret.
\end{Proof}
Finally we prove Lemma \ref{cgmhbound} used to establish the regret bound above. We introduce a new function,
\begin{align*}
h_t(X) \triangleq F_{t}(X) - F_{t}(X_t^{*}).
\end{align*} 
Also the shorthand that $h_t = h_t(X_t)$. These functions are defined conditioned on the filtration $\mathcal{F}_{t-1}$ and $f_t$. 
\begin{lemma} \label{cgmhbound}
If the parameters $\eta$ and $\gamma_t$ are chosen as stated in Theorem \ref{t:cgmregret}, such that $\eta \mathcal{G}  \sqrt{h_{t+1}} \le \mathcal{G}^2 \gamma_t^2$, the iterates $X_t$ satisfy, $h_t \le 4\mathcal{G}^2\gamma_t$.
\end{lemma}
\begin{Proof}
The functions $F_{t}$ is $1$-smooth therefore we have,
\begin{align*}
    h_t(X_{t+1}) & = F_t(X_{t+1}) - F_t(X_t^{*})= F_t(X_t + \gamma_t(\Phi(v_t) - X_t)) -F_t(X_{t}^{*})\\
    & \overset{(i)}{\le} F_t(X_t) - F_t(X_t^{*}) + \gamma_t\langle \Phi(v_t) - X_t, \nabla F_t(X_t) \rangle_{\mathcal{H}} + \frac{\gamma_t^2}{2} \lVert \Phi(v_t) - X_t \rVert_{\mathcal{H}}^2\\
    & \overset{(ii)}{\le} (1-\gamma_t)\Big(  F_t(X_t) - F_t(X_t^{*})\Big) + \gamma_t^2 \mathcal{G}^2,
\end{align*}
where $(i)$ follows by the strong convexity of $F_t$ and $(ii)$ follows as $\Phi(v_t)$ is the minimizer of $F_t(\cdot)$. By the definition of $F_{t+1}(\cdot)$ and $h_t$ we also have,
\begin{align}
    h_{t+1}(X_{t+1}) & =  F_t(X_{t+1}) - F_t(X_{t+1}^*) + \eta \langle w_{t+1}, X_{t+1} - X_{t+1}^* \rangle_{\mathcal{H}} \nonumber \\
   & \overset{(i)}{\le} F_t(X_{t+1}) -  F_t(X_{t}^*)+ \eta \langle w_{t+1},X_{t+1} - X_{t+1}^* \rangle_{\mathcal{H}} \nonumber \\
 & \overset{(ii)}{\le} h_t(X_{t+1}) + \eta \mathcal{G} \lVert X_{t+1} - X_{t+1}^{*} \rVert_{\mathcal{H}}, \label{hxbound}
\end{align}
where $(i)$ follows as $X_{t}^{*}$ is the minimizer of $F_t$ and $(ii)$ is by Cauchy-Schwartz inequality. Again by leveraging the strong convexity of $F_t$ we have, $\lVert X - X_{t+1}^{*} \rVert_{\mathcal{H}}^2 \le F_{t+1}(X) - F_{t+1}(X_{t+1}^{*}) = h_{t+1} $ which leads to the string of inequalities,
\begin{align*}
    h_{t+1}(X_{t+1}) &\le h_t(X_{t+1}) + \eta \mathcal{G}  \lVert X_{t+1} - X_{t+1}^{*} \rVert_{\mathcal{H}}  \le h_t(X_{t+1}) + \eta \mathcal{G} \sqrt{h_{t+1}(X_{t+1})}.
\end{align*}
Plugging in the bound on $h_t(X_{t+1})$ from Equation \eqref{hxbound} into the above inequality gives us the recursive relation,
\begin{align*}
    h_{t+1} &\le h_{t}(1 - \gamma_t) + \gamma_t^2 \mathcal{G}^2  + \eta \mathcal{G} \sqrt{h_{t+1}} \overset{(i)}{\le}  h_{t}(1 - \gamma_t) + 2\gamma_t^2\mathcal{G}^2,
\end{align*}
where, the last step follows by our choice of the schedule for the mixing rate $\gamma_t$ such that $\eta \mathcal{G}\sqrt{h_{t+1}} \le \mathcal{G}^2 \gamma_t^2$. We now complete the proof by an induction over $t$. 

For the base case $t=1$, we have $h_1 = F_1(X_1) - F_1(X_1^{*}) =   \lVert X_1 - X_1^{*} \rVert^{2}  \le 4\gamma_1 \mathcal{G}^2$. Thus, by the induction hypothesis for the step $t+1$ we have,
\begin{align*}
    h_{t+1} & \le h_t(1 - \gamma_t) + 2\gamma_t^2 \mathcal{G}^2\overset{(i)}{\le} 4\mathcal{G}^2(\gamma_t (1- \gamma_t)) + 2\gamma_t^2 \mathcal{G}^2=  4\mathcal{G}^2\gamma_t\left(1-\frac{\gamma_t}{2}\right)\overset{(ii)}{\le} 4\mathcal{G}^2 \gamma_{t+1},
\end{align*}
where $(i)$ follows by the upper bound on $h_t$, $(ii)$ is by the definition $\gamma_{t} = \min\left(1,\frac{2}{t^{1/2}}\right)$.
\end{Proof}
\section{Application: Posynomial Losses}
\label{sec:posynomials}
In this section we  will define a \emph{posynomial game}, by introducing posynomial losses and prove that these losses can also be viewed as kernel inner products. We will use the connection between optimizing posynomials and \emph{Geometric Programs} to prove that conditional gradient descent can be run efficiently on this family of losses. 

\begin{definition}[Monomial] A function $f: \mathbb{R}_+^d \mapsto \mathbb{R}$ defined as
\begin{align*}
f(x) = cx_1^{\alpha_1}x_2^{\alpha_2}\cdots x_d^{\alpha_d},
\end{align*}
where $c>0$ and $\alpha_i \in \mathbb{R}$, is called a monomial function.
\end{definition}
A non-negative linear combination of monomials is a posynomial.
\begin{definition}[Posynomial]
A function $f: \mathbb{R}_+^d \mapsto \mathbb{R}$ defined as
\begin{align*}
f(x) = \sum_{k=1}^m c_k x_1^{\alpha_{1k}}x_2^{\alpha_{2k}} \cdots x_d^{\alpha_{dk}},
\end{align*}
where $c_k > 0 $ and $\alpha_{i k} \in \mathbb{R}$, is called a posynomial function.
\end{definition}
Note that posynomial functions are closed under addition, multiplication and non-negative scaling. Assume the adversary at each round plays a vector of dimension $m$ with all non-negative entries, $w_t = (c_1,c_2,\cdots, c_m)$, while the player chooses a vector $x \in \mathbb{R}^d_+$. This vector is then partitioned into $m$ parts,
\begin{align*}
x = (\underbrace{x_1,x_2}_{s_1},\ldots, \underbrace{x_{d-2},x_{d-1},x_d}_{s_m}),
\end{align*}
and the feature vector is defined as 
\begin{align*}
\Phi(x) = \begin{bmatrix}
x_1^{\alpha_{1}}x_2^{\alpha_2} \\
\vdots \\
x_{d-2}^{\alpha_{d-2}}x_{d-1}^{\alpha_{d-1}}x_{d}^{\alpha_{d}}
\end{bmatrix}.
\end{align*}
Where the $i^{th}$ component of $\Phi(\cdot)$ is only a function of the $i^{th}$ partition of the coordinates $s_i$. Then the loss obtained on the evaluation of the inner product between the adversary and player action is a posynomial loss function,
\begin{align*}
\langle w_t, \Phi(x) \rangle_{\mathcal{H}} = \sum_{k=1}^m c_k x_1^{\alpha_{k1}} \cdots x_d^{\alpha_{kd}}.
\end{align*}
A number of scenarios can be modeled as a minimization/maximization problem over posynomial functions \citep[see][for a detailed list of examples]{boyd2007tutorial}. We now show that conditional gradient descent can be run efficiently over posynomial losses. If we again assume that the set of actions $\mathcal{A} = \{a \in \mathbb{R}^d : \lVert a \rVert_2 \le 1 \}$. Additionally we choose the initial action to be the solution to the optimization problem,
\begin{align*}
a_1 & = \argmin_{a \in \mathcal{A}}  \sum_{k=1}^d \Phi(a)_i. 
\end{align*}
The objective function is a posynomial subject to a posynomial inequality constraint. This is a geometric program that can be solved efficiently by changing variables and converting into a convex program \citep[Section 2.5 in][]{boyd2007tutorial}. At each round of the conditional gradient descent algorithm requires us to solve the optimization problem,
\begin{align} \label{eq:objposy}
v_t = \argmin_{a \in \mathcal{A}} \langle \eta \sum_{s=1}^{t-1} w_t + 2(X_t - \Phi(a_1)), \Phi(a) \rangle_{\mathcal{H}}.
\end{align}
Given that posynomials are closed under addition, and given our choice of $a_1$, the objective function in Equation \eqref{eq:objposy} is still a posynomial and the constraint is a posynomial inequality. This can again be cast as a geometric program that can be solved efficiently at each round.
\section{Technical Results}
\label{sec:technicalstuff}We present a version of Hoeffding's inequality \citep{hoeffding1963probability} that is used in the regret bound analysis of exponential weights.
\begin{lemma}[Hoeffding's Inequality] \label{hff}Let $\lambda>0$ and $X$ be a bounded random variable such that $\lambda X \ge -1$, then,
\begin{align*}
    \log\left(\mathbb{E}\left[ e^{-\lambda X}\right]\right) \le (e-2)\lambda^2 \mathbb{E}\left[ X^2\right] - \lambda \mathbb{E}\left[X \right],
\end{align*}
and hence
\begin{align}
     \mathbb{E}\left[ X\right] \le -\frac{1}{\lambda} \log\left(\mathbb{E}\left[ e^{-\lambda X}\right] \right) + (e-2)\lambda^2\mathbb{E}\left[ X^2\right]. \label{hoeff}
\end{align}
\end{lemma}
\begin{Proof}We look at the log of the moment generating function to get,
\begin{align*}
    \log\left(\mathbb{E}\left[\exp(-\lambda X) \right]\right) & \overset{(i)}{\le} \mathbb{E}\left[ \exp(-\lambda X)\right] - 1 \overset{(ii)}{\le} -\lambda \mathbb{E}\left[X\right] + (e-2)\lambda\mathbb{E}\left[X^2\right],
\end{align*}
where $(i)$ follows by the inequality $\log(y) \le y-1$ for all $y>0$ and $(ii)$ is by the bound $e^{-x} \le 1-x +(e-2)x^2$ for $x\ge -1$.
\end{Proof}
\subsection{John's Theorem}\label{app:Johntheorem}

We present John's theorem \citep[see][]{ball1997elementary} that we use to construct an exploration distribution.
\begin{theorem}[John's Theorem]\label{Johnstheorem} Let $\mathcal{K} \subset \mathbb{R}^d$ be a convex set, denote the ellipsoid of minimal volume containing it as,
\begin{align*}
\mathcal{E} : = \left\{x\in \mathbb{R}^d \Big\lvert (x-c)^{\top} H (x-c) \le 1  \right\}.
\end{align*}
Then there is a set $\{u_1,\ldots,u_{q} \} \subset \mathcal{E}\cap\mathcal{K}$ with $q \le d(d+1)/2 +1$ contact points and a distribution $p$ (John's distribution) on this set such that any $x \in \mathbb{R}^d$ can be written as
\begin{align*}
x = c + d\sum_{i=1}^{q}p_i\langle x-c,u_i-c \rangle_{J} (u_i - c),
\end{align*}
where $\langle \cdot, \cdot \rangle_{J}$ is the inner product for which the minimal ellipsoid is the unit ball about its center $c: \langle x,y\rangle_{J} = x^{\top} H y$ for all $x,y \in  \mathbb{R}^d$.
\end{theorem}

This shows that 
\begin{align*}
x -c & = d\sum_i p_i (u_i - c)(u_i - c)^{\top}H(x - c) \\
\iff \tilde{x} &= d \sum_{i}p_i \tilde{u}_i \tilde{u}_i^{\top}\tilde{x} \\
\iff \frac{1}{d}I_{d\times d} & = \sum_{i}p_i \tilde{u_i}\tilde{u_i}^{\top}
\end{align*}
where $\tilde{u}_i = H^{1/2}(u_i-c)$, and similarly for $\tilde{x}$. We see that for any $a,b \in \mathcal{K}$,
\begin{align*}
\tilde{a}^{\top}\mathbb{E}_{u\sim p}\left[uu^{\top} \right]\tilde{b} = \frac{1}{d}\tilde{a}^{\top}\tilde{b}. \numberthis
\end{align*}

To use this theorem, we need to perform a preprocessing of the action set $\mathcal{A}$ following a similar procedure described in Section 3 by \cite{bubeck2012towards}:

\begin{itemize}
\item First we map $\mathcal{A}$ onto the RKHS generated by the kernel $\kc$ to produce $\Phi_m(\mathcal{A})$. 
\item We assume that $\Phi_m(\mathcal{A})$ is full rank in $\mathbb{R}^m$. If not, we can redefine the feature map $\Phi_m$ as the projection onto a lower dimensional subspace.
\item Find John's ellipsoid for $Conv(\Phi_m(\mathcal{A}))$ which we denote by $\mathcal{E} = \{ x  \in \mathbb{R}^m : (x-x_0)^\top H^{-1} (x-x_0) \leq 1\}$.
\item Translate $\Phi_m(\mathcal{A})$ by $x_0$. In other words, assume that $\Phi_m(\mathcal{A})$ is centered around $x_0 = 0$ and define the inner product $\langle x , y \rangle_J = x^\top H y$. 
\item We now play on the set $\Phi_m^{J}(\mathcal{A}) := H^{-1} \Phi_m(\mathcal{A})$ in $\mathbb{R}^m$. Let the loss of playing an action $H^{-1}\Phi_m(a) \in  \Phi_m^{J}(\mathcal{A})$ when the adversary plays $z$ be $\langle H^{-1}\Phi_m(a), z \rangle_J = \Phi_m(a)^\top z$. 
\item The contact points, $u_1,\ldots,u_q$ are in $\Phi_m^{J}(\mathcal{A})$ and are valid points to play. We now use $p$ -- John's distribution -- to be the exploration distribution.
\end{itemize}
Mimicking \cite{bubeck2012towards} it can be shown that Algorithm \ref{a:expweightskernels} works with a generic dot product and that all the steps in the regret bound in Appendix \ref{app:regretboundbandit} go through.

\section{Experiments}
\label{sec:experiments}
We perform an empirical study of our algorithms in both the full information and the bandit settings and demonstrate their practicality. In the full information setting we conducted experiments with quadratic losses using exponential weights. We also plot the performance of exponential weights algorithm on Gaussian losses. In the bandit feedback setting we again study quadratic and Gaussian losses.
\subsubsection*{Full information}
\begin{figure}[h]
  \centering
  \begin{minipage}[b]{0.445\textwidth}
    \includegraphics[width=\textwidth]{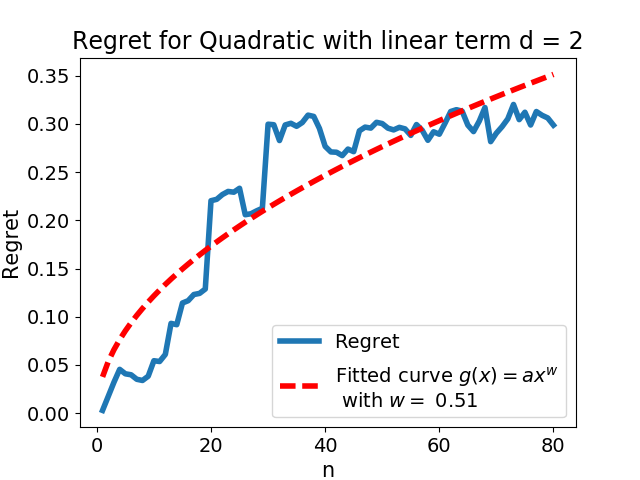}
    \caption{Quadratic with linear term Full Information.}
  \end{minipage}
  \hfill
  \begin{minipage}[b]{0.445\textwidth}
    \includegraphics[width=\textwidth]{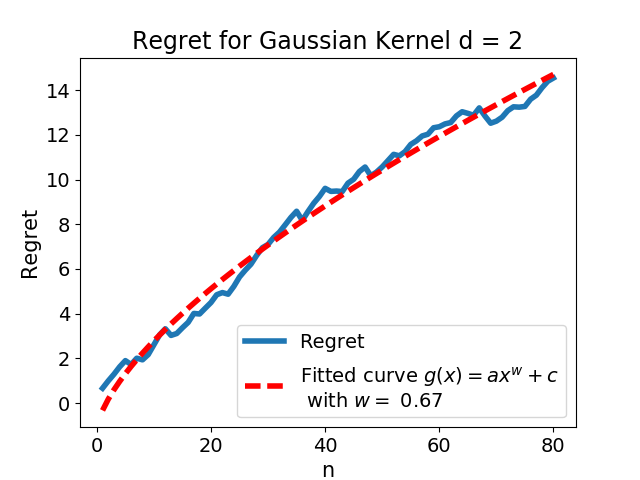}
    \caption{Gaussian Losses Full Information.}
  \end{minipage} 
\end{figure}
Exponential weights requires us to sample from a distribution of the form $p(x) \propto \exp( \mu \sum_{i=1}^t \ke(x, w_i) )$. In general sampling from these distributions is possibly intractable, however they present good empirical performance.  The following plot shows a diffusion MCMC algorithm sampling from a distribution proportional to $\exp\left ( -\eta\sum_{i=1}^t \ke(x, z_i)\right)$ where $\ke$ is the Gaussian kernel, $\eta = 10$, and $x$ is restricted to an $\ell_2$ ball of radius $10$. 
In practice using exponential weights in the full information setting and sampling using a diffusion MCMC algorithm yields sublinear regret profiles and tractable sampling even for Gaussian losses. We ran experiments generating random loss sequences and we plot the average regret over 60 runs of the algorithm. 
\subsubsection*{Bandits Experiments}
\begin{figure}[h]
  \centering
  \begin{minipage}[b]{0.445\textwidth}
    \includegraphics[width=\textwidth]{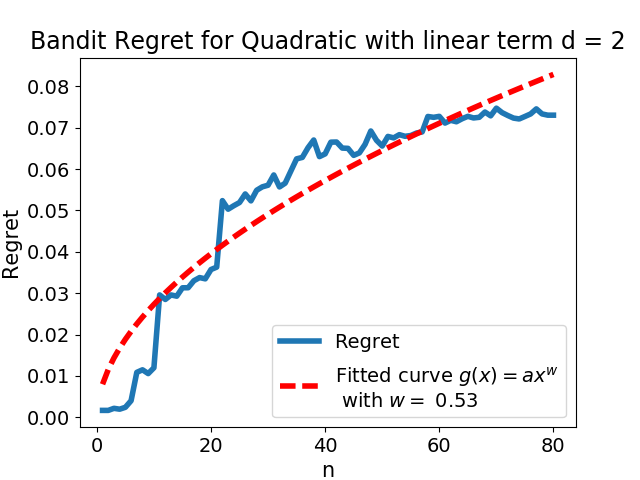}
    \caption{Quadratics with linear term Bandit Feedback.}
  \end{minipage}
  \hfill
  \begin{minipage}[b]{0.445\textwidth}
    \includegraphics[width=\textwidth]{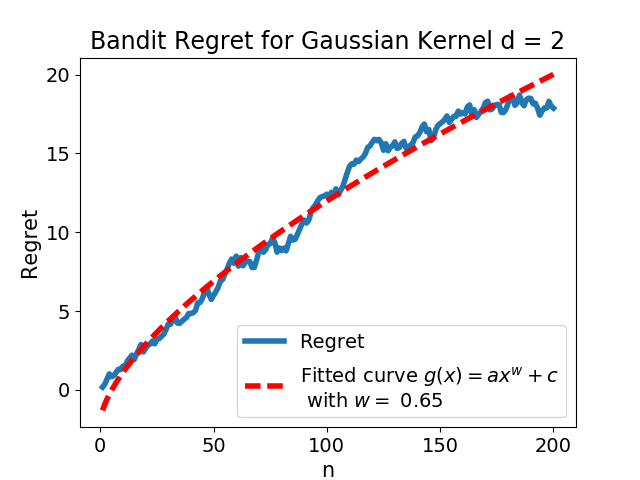}
    \caption{Gaussian Losses Bandit Feedback.}
  \end{minipage}
\end{figure}
The kernel exponential weights algorithm presents also a sublinear regret profile. The Gaussian experiments involved the construction of the finite dimensional kernel $\ke_m$ by kernel PCA. 

\end{document}